\documentclass[letterpaper, 10 pt, conference]{ieeeconf}  % Comment this line out if you need a4paper

\IEEEoverridecommandlockouts                              % This command is only needed if 
\usepackage{graphicx}
\usepackage{amsmath}
\usepackage{paralist}
\usepackage{subcaption}
\usepackage{siunitx}
\usepackage{wrapfig}
\usepackage{dsfont}
\let\labelindent\relax
\usepackage{enumitem}
\setlist{nolistsep}
\usepackage{hyperref}
\usepackage[linesnumbered,algoruled,boxed,lined]{algorithm2e}
\hypersetup{
    colorlinks=true,
    linkcolor=black,
    filecolor=magenta,      
    urlcolor=blue,
    pdftitle={How Does It Feel? Self-Supervised Costmap Learning for Off-Road Vehicle Traversability},
    pdfpagemode=FullScreen,
    }

\usepackage{cite}

\title{\LARGE \bf
How Does It Feel? Self-Supervised Costmap Learning for Off-Road Vehicle Traversability
}

\author{Mateo Guaman Castro$^{1}$, Samuel Triest$^{1}$, Wenshan Wang$^{1}$, Jason M. Gregory$^{2}$,\\ Felix Sanchez$^{3}$, John G. Rogers III$^{2}$, and Sebastian Scherer$^{1}$% <-this % stops a space
\thanks{*Research was sponsored by the Army Research Laboratory and was accomplished under Cooperative Agreement number W911NF-21-2-0152. The views and conclusions contained in this document are those of the authors and should not be interpreted as representing the official policies, either expressed or implied, of the Army Research Laboratory or the U.S. Government. The U.S. Government is authorized to reproduce and distribute reprints for Government purposes notwithstanding any copyright notation herein.}% %W911NF2120152.}% <-this % stops a space
\thanks{$^{1}$ Robotics Institute, Carnegie Mellon University, Pittsburgh, PA, USA. \{mguamanc, striest,wenshanw,basti\}@andrew.cmu.edu}%
\thanks{$^{2}$ DEVCOM Army Research Laboratory, Adelphi, MD, USA. \{jason.m.gregory1.civ, john.g.rogers59.civ\}@army.mil}%
\thanks{$^{3}$Booz Allen Hamilton,
        McLean, VA 22102, USA
        sanchez\_felix@bah.com}%
}

\begin{document}

\maketitle
\thispagestyle{empty}
\pagestyle{empty}

%%%%%%%%%%%%%%%%%%%%%%%%%%%%%%%%%%%%%%%%%%%%%%%%%%%%%%%%%%%%%%%%%%%%%%%%%%%%%%%%
\begin{abstract}

    Estimating terrain traversability in off-road environments requires reasoning about complex interaction dynamics between the robot and these terrains. However, it is challenging to create informative labels to learn a model in a supervised manner for these interactions. We propose a method that learns to predict traversability costmaps by combining exteroceptive environmental information with proprioceptive terrain interaction feedback in a self-supervised manner. Additionally, we propose a novel way of incorporating robot velocity into the costmap prediction pipeline. We validate our method in multiple short and large-scale navigation tasks on challenging off-road terrains using two different large, all-terrain robots. Our short-scale navigation results show that using our learned costmaps leads to overall smoother navigation, and provides the robot with a more fine-grained understanding of the robot-terrain interactions. Our large-scale navigation trials show that we can reduce the number of interventions by up to 57\% compared to an occupancy-based navigation baseline in challenging off-road courses ranging from 400 m to 3150 m. Appendix and full experiment videos can be found in our website: \url{https://mateoguaman.github.io/hdif}.

\end{abstract}

%%%%%%%%%%%%%%%%%%%%%%%%%%%%%%%%%%%%%%%%%%%%%%%%%%%%%%%%%%%%%%%%%%%%%%%%%%%%%%%%
\section{Introduction}
\label{sec:introduction}

    Outdoor, unstructured environments are challenging for robots to navigate. Rough interactions with terrain can result in a number of undesirable effects, such as rider discomfort, error in state estimation, or even failure of robot components. Unfortunately, it can be challenging to predict these interactions a priori from exteroceptive information alone. Certain characteristics of the terrain, such as slope, irregularities in height, the deformability of the ground surface, and the compliance of the objects on the ground, affect the dynamics of the robot as it traverses over these features. While these terrain characteristics can be sensed by proprioceptive sensors like Inertial Measurement Units (IMUs) and wheel encoders, these modalities require direct contact with the terrain itself. Additionally, the robot's interaction with the ground leads to dynamic forces which are proportional to velocity and suspension characteristics. In order to \textit{feel} what navigating over some terrain at some velocity is like, we argue that the robot must actually traverse over it.
    
    \begin{figure}[t]
        \centering
        \includegraphics[width=\linewidth]{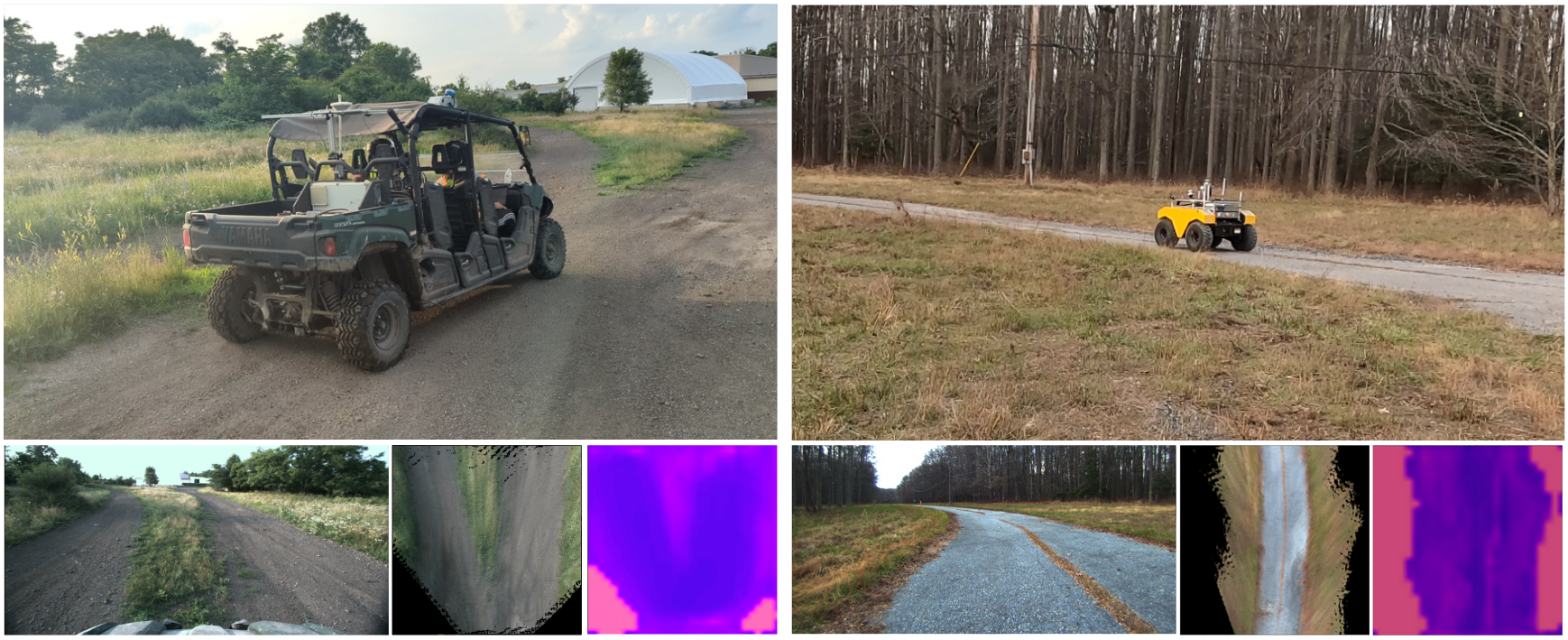}
        \caption{We present a costmap learning method for off-road navigation. We demonstrate the efficacy of our method on a large ATV (left), and on a Warthog UGV robot (right).}
        \label{fig:main_fig}
        \vspace{-0.15in}
    \end{figure}
    
    Previous approaches for off-road traversability have focused on representing exteroceptive information as occupancy maps \cite{fu2021coupling, fan2021step}, or learning semantic classifiers from labeled data to map different terrains to different costs in a costmap \cite{maturana2018real, kim2018season}. Yet, this abstracts away all the nuance of the interactions between the robot and different terrain types. Under an occupancy-based paradigm, concrete, sand, and mud would be equally traversable, whereas tall rocks, grass, and bushes would be equally \textit{non}-traversable. In reality, specific instances of a class may have varying degrees of traversability (e.g. some bushes are traversable but not all). 
    
    Other approaches have characterized terrain roughness directly from geometric features \cite{lalonde2006natural, krusi2017driving, gennery1999traversability}. Yet, what we are really interested in capturing is roughness as the vehicle experienced it, rather than capturing the appearance or geometry of roughness. For instance, a point cloud of tall grass might appear rough, but traversing over this grass could still lead to smooth navigation if the terrain under the grass is smooth. Finally, other learning-based methods learn predictive models or direct control policies for off-road navigation \cite{kahn2021badgr}. However, many of these do not take into account robot dynamics, which are fundamental in scenarios where the state of the robot, such as its velocity, can lead to a wide range of behaviors. For instance, speeding up before driving over a bump in the terrain can lead to jumping, whereas slowing down will usually result in smoother navigation. 
    
    In this paper, we propose a self-supervised method that predicts costmaps that reflect nuanced terrain interaction properties relevant to ground navigation. Motivated by examples in legged locomotion \cite{wellhausen2019should}, we approach this problem by learning a mapping from rich exteroceptive information and robot velocity to a continuous traversability cost derived from IMU data. We propose a learning architecture which combines a CNN backbone to process high-dimensional exteroceptive information with a feed-forward network that takes in a Fourier parameterization of the low-dimensional velocity information, inspired by recent advances in implicit representation learning \cite{mildenhall2020nerf, tancik2020fourier}. 
    
    Our main contributions are: a) learning to aggregate visual, geometric, and velocity information directly into a \textit{continuous-valued} costmap, without human-annotated labels, and b) demonstrating ease of integration into traditional navigation stacks to improve navigation performance, due to our choice of using a top-down metric representation. In the rest of this paper, we present our contributions in more detail as follows:
    
    \begin{itemize}[leftmargin=*]
        \item We present an IMU-derived traversability cost that can be used as a self-supervised pseudo ground-truth for training.
        \item We demonstrate a novel way to combine low-dimensional dynamics information with high-dimensional visual features through Fourier feature mapping \cite{tancik2020fourier}. 
        \item We propose a system that produces continuous-valued learned costmaps through a combination of visual and geometric information, and robot velocity.
        \item We validate our method on outdoor navigation tasks using two different ground robots.
    \end{itemize}

%===============================================================================

\begin{figure*}[ht]
    \centering
    \includegraphics[width=0.85\linewidth]{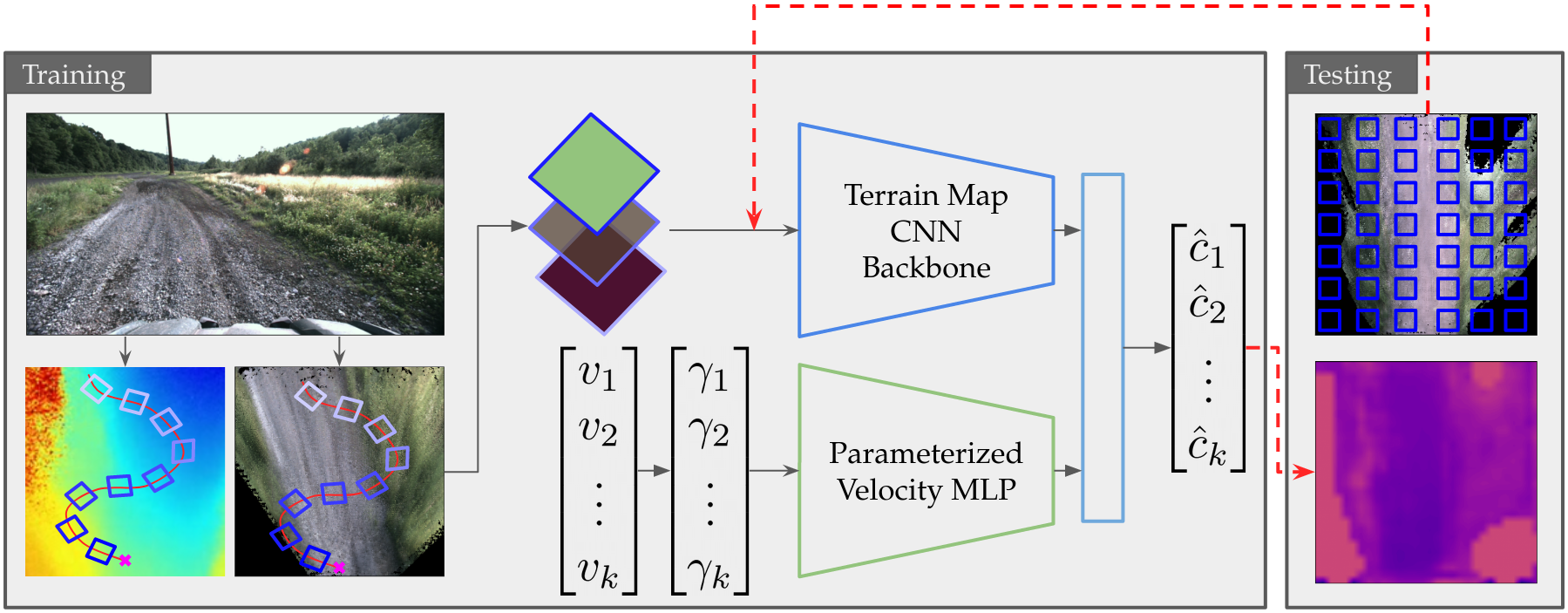}
    \caption{\textbf{System Overview}: During training, the network takes in patches cropped from a top-down colored map and height map along the driving trajectory, as well as the parameterized velocity corresponding to each patch. The network predicts a traversability cost for each patch, supervised by a pseudo ground-truth cost generated from IMU data. During testing, the whole map is subsampled into small patches, which are fed into the network to generate a dense, continuous costmap.}
    \label{fig:costmap}
    \vspace{-0.15in}
\end{figure*}

\section{Related Work}
\label{sec:related}

    There exists over a decade of work in learning methods for off-road traversability \cite{howard2006towards, angelova2007learning, hadsell2009learning, dupont2008frequency, konolige2009mapping, bajracharya2009autonomous, stavens2006self}, much of which can be traced to the DARPA LAGR \cite{jackel2006darpa} and Grand Challenge \cite{thrun2006stanley} programs. 
    % These methods applied lightweight machine learning techniques \cite{angelova2007learning, howard2006towards, konolige2009mapping, bajracharya2009autonomous, stavens2006self}, and even early convolutional neural networks (CNNs) \cite{hadsell2009learning}, to classify terrain types according to their traversability based on visual and geometric appearance.
    These methods learned lightweight terrain traversability classifiers based on visual and geometric appearance \cite{angelova2007learning, howard2006towards, konolige2009mapping, bajracharya2009autonomous, stavens2006self, hadsell2009learning}.
    Other approaches directly estimate a terrain roughness score by analyzing planarity \cite{krusi2017driving} or eigenvalues of the terrain point cloud \cite{lalonde2006natural}. 
    % Stavens et al. \cite{stavens2006self} learn a terrain roughness estimator from geometric data, and they find a strong correlation between vehicle velocity and roughness estimated from $z$-axis linear acceleration. Similarly, Dupont et al. \cite{dupont2008frequency} show that the frequency response of IMU data can be used to classify different terrain types. 
    % The importance of conditioning vibration-based traversability analysis methods on robot velocity  has been recently highlighted in \cite{borges2022survey}.

    Much of the recent literature on learning for off-road traversability estimation has focused on semantic segmentation of visual data into discrete classes \cite{maturana2018real, long2015fully}. 
    % Maturana et al. \cite{maturana2018real} and Kim et al. \cite{kim2018season} learn semantic segmentation of off-road terrain classes--such as grass, gravel, and trees--using a Fully Convolutional Network (FCN) backbone \cite{long2015fully}, and combine it with geometric lidar data to generate a metric map that can be used for path planning for a large ground robot. 
    However, it is not immediately obvious how to map human semantic classes into costs that can be directly used for path planning and control, which often results in hard-coded mappings. More recent work by Shaban et al. \cite{shaban2022semantic} aims to alleviate the need for explicit mappings from human semantic classes to costs by directly learning discrete traversability classes, such as low-cost and high-cost, in a metric space from geometric data. Yet, these approaches require a large amount of labeled semantic data for training, and lack expressiveness given the limited number of traversability classes.

    Recently, other works have looked into learning predictive models, control policies, and risk-aware costmaps directly from visual and multimodal inputs for navigation in challenging off-road environments. 
    While BADGR-based methods \cite{kahn2021badgr, shah2022rapid} learn a \textit{boolean} predictor for whether a specific sequence of actions will lead to a bumpy trajectory, we learn a continuous value for traversability that is aggregated into a costmap that can be used directly to optimize trajectories, without the need to query the network for every sample of our MPPI optimizer \cite{williams2017information}. 
    Triest et al. \cite{triest2022tartandrive} learn a neural network-based dynamics model for a large ATV vehicle and explore the use of different types of multimodal data as input to their neural network.
    Sivaprakasam et al. \cite{sivaprakasam2021improving} learn a dynamics model in simulation to derive a dynamics-aware cost function for downstream planning tasks. 
    Fan et al. \cite{fan2021learning} learn a traversability risk costmap from lidar data, and Cai et al. \cite{cai2022risk} learn a speed distribution map that is converted into a costmap using a conditional value at risk (CVaR) formulation.
    Triest et al. \cite{triest2023learning} learn CVaR-based uncertainty-aware traversability costmaps from lidar data using inverse reinforcement learning.
    
    % Our approach is heavily influenced by the legged locomotion work by \cite{wellhausen2019should}, where they learn a "ground reaction score" (GRS) directly from front-facing images captured by a legged robot to learn a continuous-valued segmentation of the image based on GRS in a self-supervised fashion. 
    Most recently, traversability estimation for off-road robots has shifted towards learning continuous costmaps in a self-supervised manner with IMU signals as learning targets, with these methods learning from RGB data \cite{wellhausen2019should, sathyamoorthy2022terrapn, yao2022rca}, or point clouds \cite{waibel2022rough}, and \cite{sathyamoorthy2022terrapn, waibel2022rough} conditioning on robot speed similar to our approach. We use both RGB and point cloud data in our approach to generate high-resolution, continuous costmaps, and we demonstrate our approach in large-scale, challenging off-road courses at much higher speeds on two different large robot platforms.

%===============================================================================

\section{Costmap Learning}
\label{sec:costmap}

	We now introduce our self-supervised costmap learning method, which associates a proprioception-derived cost to high-dimensional visual and geometric data, and robot velocity. In the following section, we will describe our costmap learning pipeline (Figure \ref{fig:costmap}), which consists of:
	\begin{enumerate}[leftmargin=*]
	    \item Derivation of a cost function from proprioception,
	    \item Extraction of a map-based representation from visual and geometric data,
	    \item Representation of velocity as an input to our model, and
	    \item Training a neural network to predict traversability cost.
	\end{enumerate}

\subsection{Pseudo-Ground Truth Traversability Cost}
\label{sec:traversability}
	
	We aim to learn a continuous, normalized traversability cost that describes the interactions between the ground and the robot, and which can be directly used for path planning. As shown in previous work \cite{wellhausen2019should, stavens2006self, dupont2008frequency}, linear acceleration in the $z$ axis, as well as its frequency response, capture traversability properties of the environment that not only depend on the characteristics of the ground, but also on the speed of the robot. To obtain a single scalar value that generally describes traversability properties of the terrain, such as the roughness, bumpiness, and deformability, we use the bandpower of the IMU linear acceleration of the robot in the $z$ axis:
	
    \vspace{-0.15in}
	\begin{equation}
	    \hat{c} = \int_{f_\text{min}}^{f_\text{max}} S^{W}_{a_z}(f) \,d f,
	\end{equation}
	
	where $\hat{c}$ is our estimated traversability cost, $S^{W}_{a_z}$ is the power spectral density (PSD) of the linear acceleration $a_z$ in the $z$ axis, calculated using Welch's method \cite{welch1967use}, , and $f_{\text{min}}$ and $f_{\text{max}}$ describe the frequency band used to compute bandpower. We use a frequency range of 1-30 Hz, since this range highly correlates with human-labeled roughness scores, and normalize based on data statistics from recorded trajectories, as detailed in the Appendix.

\subsection{Mapping}
\label{sec:mapping}

    We represent the exteroceptive information about the environment in bird's eye view (BEV), which allows us to aggregate visual and geometric information in the same space, which we refer to as the ``local map." 
    We use a stereo matching network \cite{chang2018pyramid} to obtain a disparity image, from which we estimate the camera odometry using TartanVO \cite{wang2020tartanvo}, a learning-based visual odometry algorithm. 
    We use this odometry and the RGB data to register and colorize a dense point cloud which we then project into a BEV local map. 
    The local map consists of a stacked RGB map, containing the average RGB value of each cell, and a height map, containing the minimum, maximum, mean, and standard deviation of the height of the points in each cell, ignoring all points 2 meters above the ground surface to deal with overhangs. 
    Additionally, we include a boolean mask that describes areas in the local map for which we have no information, either due to occlusions or limited field of view of the sensors.

\subsection{Velocity Parameterization}
\label{sec:fourier}

    At high speeds, rough terrain leads to higher shock sensed by the vehicle, proportional to its speed, as explored in \cite{stavens2006self}. 
    % An expert human driver will likely slow down before reaching rough terrain areas in order to reduce bumpiness.
    We obtain velocity-conditioned costmaps by including robot velocity as an input to our network.
    
    In order to balance the high-dimensional local map input and the low-dimensional velocity input, we use Fourier feature mapping \cite{tancik2020fourier} on the robot's velocity. Recent advances in implicit neural representations have shown that mapping a low-dimensional vector (or scalar) to a higher dimensional representation using Fourier features can modulate the spectral bias of MLPs (which is usually biased towards low-frequency functions \cite{rahaman2019spectral}) towards higher frequencies by adjusting the scale of the Fourier frequencies. 
    Intuitively, we hypothesize that this parameterization lets the netwrok learn a function that more readily adjusts to subtle changes in velocity input, and prevents the network's predictions from being dominated by the high-dimensional 2D inputs. 
    We use the following parameterization:
    
    \vspace{-0.05in}
    \begin{equation} \label{eq:ffm}
        \gamma(v) = \begin{bmatrix} \cos(2\pi b_1 v)\\ \sin(2\pi b_1 v)\\ \vdots \\ \cos(2\pi b_m v) \\ \sin(2\pi b_m v) \end{bmatrix}
    \end{equation}
    
    In Equation \ref{eq:ffm}, $v$ corresponds to the norm of the 3D velocity vector, $b_i \sim \mathcal{N}(0, \sigma^2)$ are sampled from a Gaussian distribution with tunable scale $\sigma$, and $m$ corresponds to the number of frequencies used to map the scalar velocity value into a $2m$ dimensional vector.
	
\subsection{Costmap Learning}
\label{sec:learning}

    Our costmap learning pipeline consists of three parts: a) obtaining local map patches from robot trajectories, b) training a network to predict traversability costs from a set of patches and associated pseudo ground-truth labels, and c) populating a cost map using the trained network at test time. 
    
    \textbf{Local Map Patches}: We extract 2x2 meter patches (roughly the robot footprint) of the local map corresponding to the parts of the environment that the robot traversed over during the dataset generation. Since the patch under the robot is not observable from a front-facing view, we first register all the local maps into an aggregated map. We use the robot odometry to locate and extract the patch in the global map at a given 2D position and orientation. At each of these positions, we use a sliding window of the last one second of IMU linear acceleration data to obtain a pseudo ground-truth traversability cost as described in Section \ref{sec:traversability}. We also record the velocity at these positions for training.

    \textbf{Cost Learning}: We train a deep neural network $f_\theta(P, v)$ parameterized by weights $\theta$, as shown in Figure \ref{fig:costmap}, that takes in as input local map terrain patches $P \in \mathds{P}$ and corresponding Fourier-parameterized velocities $\gamma(v), \gamma: \mathds{R}^{+} \rightarrow \mathds{R}^{2m}$, and predicts the traversability cost of each patch $\hat{c} = f_\theta(P, \gamma), f_\theta: \mathds{P} \times \mathds{R}^{2m} \rightarrow \mathds{R}^{+}$. We use a ResNet18 \cite{he2016deep} backbone to extract features from the patches, and a 3-layer MLP to extract features from the parameterized velocity. Finally, we concatenate these features and pass them through a fully-connected layer with sigmoid activation to obtain a normalized scalar value representing the learned traversability cost. We train this network using a Mean Squared Error (MSE) loss between the predicted costs and the pseudo ground-truth values using the Adam \cite{kingma2014adam} optimizer.

    \textbf{Costmap Prediction}: We produce costmaps at test time by taking the current local map in front of the robot, extracting patches at uniformly sampled positions (with the same orientation as the robot's current orientation), and passing them into the network. We then reshape each of the cost predictions into a costmap that corresponds to the original local map. We find that it is important to add a stride in the sampling process to allow the patch cost querying through the network to run in real time. In our experiments, we subsample the local map with a stride of 0.2m, and upsample the reshaped predicted cost values back to the shape of the local map, which allows us to produce costmaps at  7-8 Hz on an onboard NVIDIA GeForce RTX 3080 Laptop GPU.

%===============================================================================

\section{Experimental Results}
\label{sec:experimental}

\subsection{Training Data}
\label{sec:data}

    To train our network, we use TartanDrive \cite{triest2022tartandrive}, a large-scale off-road dataset containing roughly 5 hours of rough terrain traversal using a commercial ATV with a sensor suite. 
    We use the stereo images in the dataset to obtain dense point clouds using TartanVO \cite{wang2020tartanvo} (section \ref{sec:mapping}), as well as the IMU and odometry data to obtain traversability costs and velocities, respectively. 
    In order to effectively train our network, we find it necessary to augment and balance the data with respect to the pseudo ground-truth traversability cost. 
    We enforce a 2:1 ratio of high to low cost frames, resulting in 15K training frames, and 3K validation frames.
    Additionally, we fine-tune the base model with 9.5K training frames and 1.3K validation frames collected on the Warthog platform for our Warthog experiments.

\subsection{Navigation Stack}
\label{sec:navigation}

% We validate our learned costmaps in off-road navigation tasks, where the goal is to navigate from the current position of the robot to a target location. 
We validate our learned costmaps in off-road navigation tasks, where the goal is to navigate to a target location. 
For state estimation, we use Super Odometry \cite{zhao2021super} on the commercial ATV and a pose graph-based SLAM system on the Warthog \cite{gregory2016application}. For path planning and control, we use model predictive path integral control (MPPI) \cite{williams2017information}. We plan through a kinematic bicycle model with actuator limits on both the ATV and the Warthog. In order to obtain costs that can be used for the MPPI optimization objective, we query the learned costmap via Equation \ref{equation:cost_fn}. This cost function queries the costmap for each state-action pair in the trajectory $\tau$ and sums it with a weighted Euclidean distance between the final state and the goal $g$ (where $p(s)$ extracts the x-y position of state $s$). Since our learned costmap only learns costs for parts of the terrain it has driven over, it will not know what cost to assign to obstacles that the robot is incapable of traversing over. We alleviate this by composing our learned costmap with a lethal height costmap for obstacle avoidance (with a high threshold), resulting in costmap $J_{map}$. $K_g$ is found empirically.

\begin{equation}
        \label{equation:cost_fn}
        J(\tau) = \sum^{T}_{t=0} [J_{map}(s_t, a_t)] + K_g ||(p(s_T) - g)||_2
    \end{equation}

\subsection{Robot Platforms}
\label{sec:robot_platforms}
We demonstrate our system on two different ground robots autonomously operating at 3 m/s: a large all-terrain autonomous commercial vehicle (ATV), and a Clearpath Warthog robot \cite{clearpath_robotics_2021}, a large unmanned ground vehicle (UGV). For more hardware details, please see our Appendix. 

%===============================================================================

\subsection{Results}
\label{sec:results}

In our experiments, we are interested in answering the following questions:

\begin{enumerate}[leftmargin=*]
    \item Do our learned costmaps capture more nuance than the baseline lethal-height costmaps?
    \item How much of an effect does velocity have in our predicted costmaps?
    \item Do we obtain more intuitive behaviors and better navigation performance using learned costmaps inside a full navigation stack?
    \item How well does our method transfer to other robots?
\end{enumerate}

\begin{figure}[t]
    \centering
    \vspace{0.05in}
    \includegraphics[width=\linewidth]{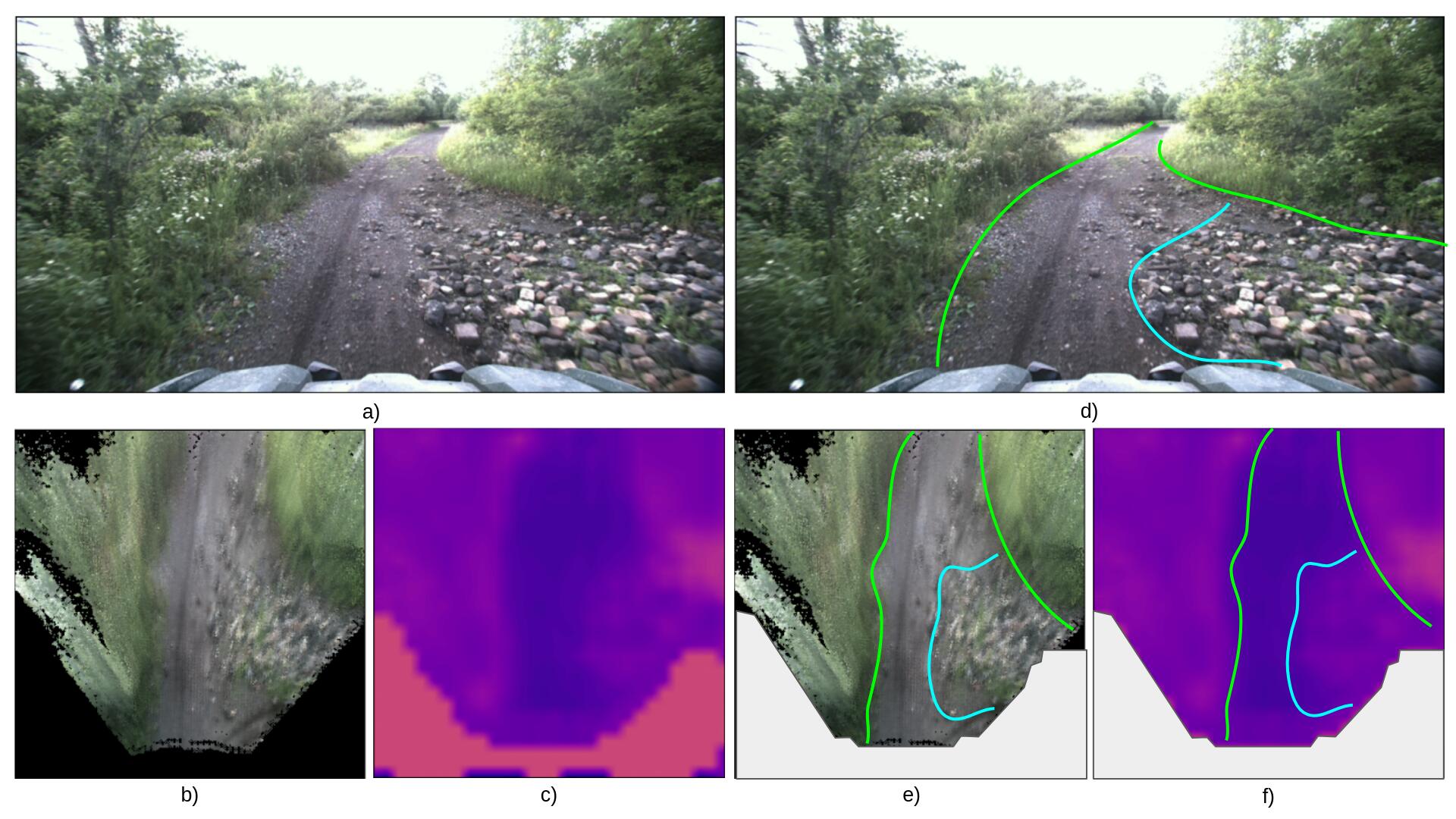}
    \caption{Learned Traversability Costmaps: a), b), and c) show the robot's front-facing view, the corresponding RGB map, and the predicted costmap respectively. d), e), and f) show a manually annotated version of a), b), and c) for easier visualization of the different traversability properties that our costmap captures. We set the unknown regions to have a cost of 0.5 as shown in c), which we mask out in f).}
    \label{fig:nuanced_costmaps}
    \vspace{-0.25in}
\end{figure}

\textbf{Do learned costmaps show more nuance?} To observe trends in the learned costmap, we tele-operated the ATV around our test site and visually analyzed the predicted costmaps in different environments. We observe that our learned costmaps are able to estimate different costs for terrains of the same height but with different traversability properties. 
% In Figure \ref{fig:nuanced_costmaps}, we show a scenario with gravel, smooth dirt, and vegetation, where our costmap predicts different costs that account for how traversing those terrains actually feels.
In Figure \ref{fig:nuanced_costmaps}, we show a scenario with gravel, smooth dirt, and vegetation, where our costmap predicts different costs for these different terrains.

We notice some particular trends in our experiments. Transitions in texture from one terrain to another are usually discernible in our costmaps. Terrains covered in grass exhibit a higher cost than smooth dirt paths. Finally, terrains with higher frequency textures, such as large patches of gravel, are predicted to have higher costs than smooth terrains. 

\begin{table}
    \centering
    \vspace{0.05in}
    \begin{tabular}{|c|c|}
    \hline
        Roughness Estimator & Val. Loss ($\times10^{-2}$) \\ \hline
        $\sigma_{z} \text{\cite{fankhauser2018robust}} + v$ & 6.08 \\ \hline
        $\sigma_{z} \text{\cite{fankhauser2018robust}} + \text{RGB} + v$ & 5.92\\ \hline
        Ours & \textbf{4.82} \\ \hline
    \end{tabular}
    \caption{Comparison to other appearance-based roughness estimators. Our method outperforms both a geometric, as well as a geometric and visual roughness estimators.}
    \vspace{-0.25in}
    \label{tab:roughness_appearance}
\end{table}

\textbf{Does velocity affect the predicted costmaps?}
We evaluate whether adding velocity as an input to our network improves the loss achieved in the validation set. We analyze three different models:
\begin{enumerate}[leftmargin=*]
    \item \textbf{patch-model}: just the ResNet backbone for patch feature extraction.
    \item \textbf{patch-vel-model}: combines the ResNet backbone with an MLP to processes normalized velocity.
    \item \textbf{patch-Fourier-vel-model}: combines the ResNet backbone with an MLP that processes Fourier-parameterized normalized velocity.
\end{enumerate}

Training results for all three models and ablations are shown for five random seeds in the Appendix. We find that adding velocity as an input to the network leads to better performance than using local map patches alone, and that the \texttt{patch-Fourier-vel-model} performs best as measured by the validation loss. Our ablations show that using both RGB and height statistics performs slightly better than using just RGB, and that the scale and number of frequencies used for Fourier parameter mapping of velocity do not make much of a difference.

We compare our method with two appearance-based roughness baselines to evaluate whether our network captures more accurately the robot-terrain interactions. The first baseline characterizes roughness as the standard deviation of height in a given patch  \cite{fankhauser2018robust}. We extend this metric with the average RGB values in a patch for our second baseline. For a fair comparison, we add normalized velocity as an input to both baselines. We learn the weights for a linear combination of these inputs through logistic regression. We compare with our best learned model in the validation set (Table \ref{tab:roughness_appearance}).

We also evaluate the effect of robot speed in the predicted costmaps in a real-robot experiment. In this experiment, we command the ATV different velocities and aggregate the average predicted cost over a straight 200 m trail with similar terrain characteristics throughout. Additionally, we integrate the sum of costs of the entire costmap (energy) along the trajectory. 
We summarize the results in Table \ref{tab:robot_velocity}, and show the resulting costmaps in the Appendix.
We find that traversability cost and overall costmap energy generally increase as robot speed increases.

\begin{table}
    \centering
    \vspace{0.05in}
    \begin{tabular}{|c|c|c|c|}
    \hline
        Velocity & Pred. Traj. Cost & GT Traj. Cost & Costmap Energy \\ \hline
        3 m/s & 0.093 & 0.038 & 0.185 \\ \hline
        5 m/s & 0.163 & 0.055 & 0.250 \\ \hline
        7 m/s & 0.201 & 0.084 & 0.289 \\ \hline
        10 m/s & 0.195 & 0.106 & 0.266 \\ \hline
    \end{tabular}
    \caption{Effect of velocity on the predicted costmap in a real-robot experiment. Higher speeds generally result in higher costs, both on the robot's trajectory and the costmap as a whole. Note that while the IMU-based ground truth cost increases monotonically as velocity increases, the predicted cost stops increasing after 7 m/s likely due to lack of enough training data at higher speeds.}
    \label{tab:robot_velocity}
\end{table}

\begin{figure}[ht]
    \centering
    \vspace{-0.10in}
    \includegraphics[width=\linewidth]{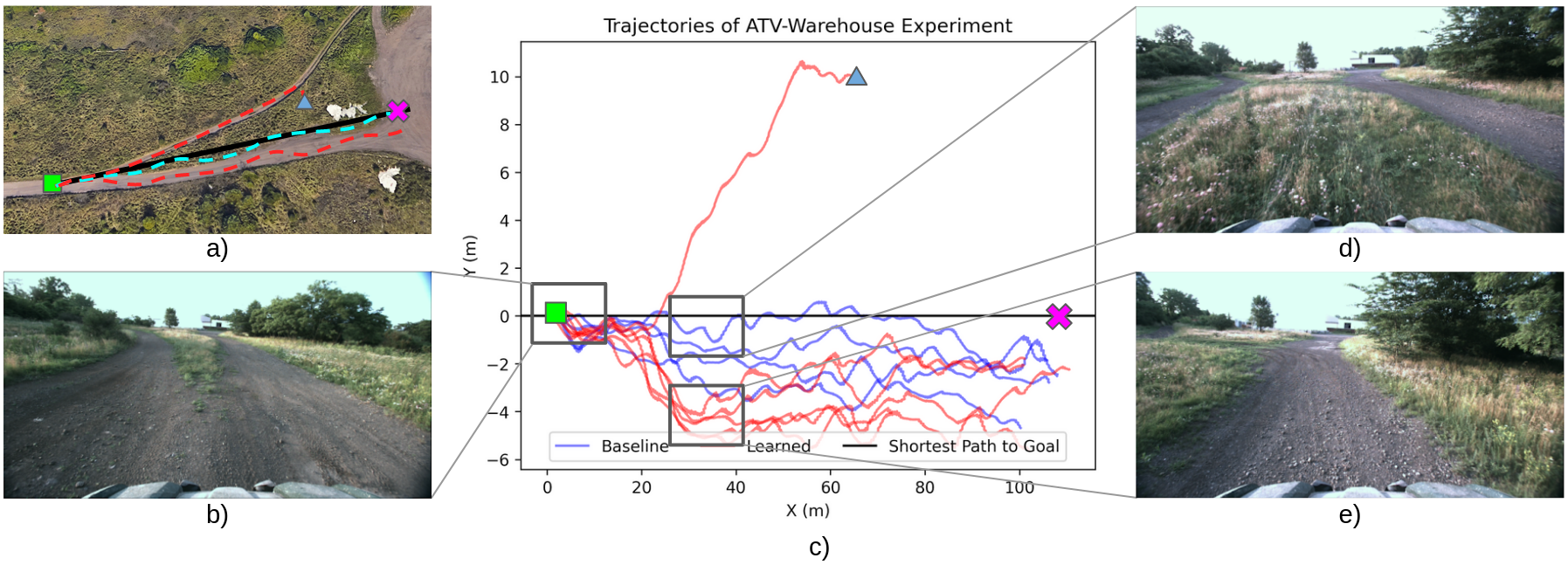}
    \caption{Short-scale navigation experiment (ATV-Warehouse). When using our costmap, the robot deviates from the straight path to the goal and chooses the smoother dirt path (e) over the patch of vegetation (d) to get to the goal. a) Sketch of the trajectories taken when using the baseline stack (blue) and our stack (red). The green square is the start position, and the pink cross is the goal position. One trajectory ends in an intervention (triangle) due to the robot taking an equally-smooth dirt path to avoid the grass. b), d), and e) show front-facing views at those points in the trajectory. c) shows all trajectories, as well as the direct path to the goal (black).}
    \label{fig:nav_experiments}
    % \vspace{-0.2in}
\end{figure}

\begin{figure}[t!]
    \centering
    % \vspace{-0.15in}
    \includegraphics[width=\linewidth]{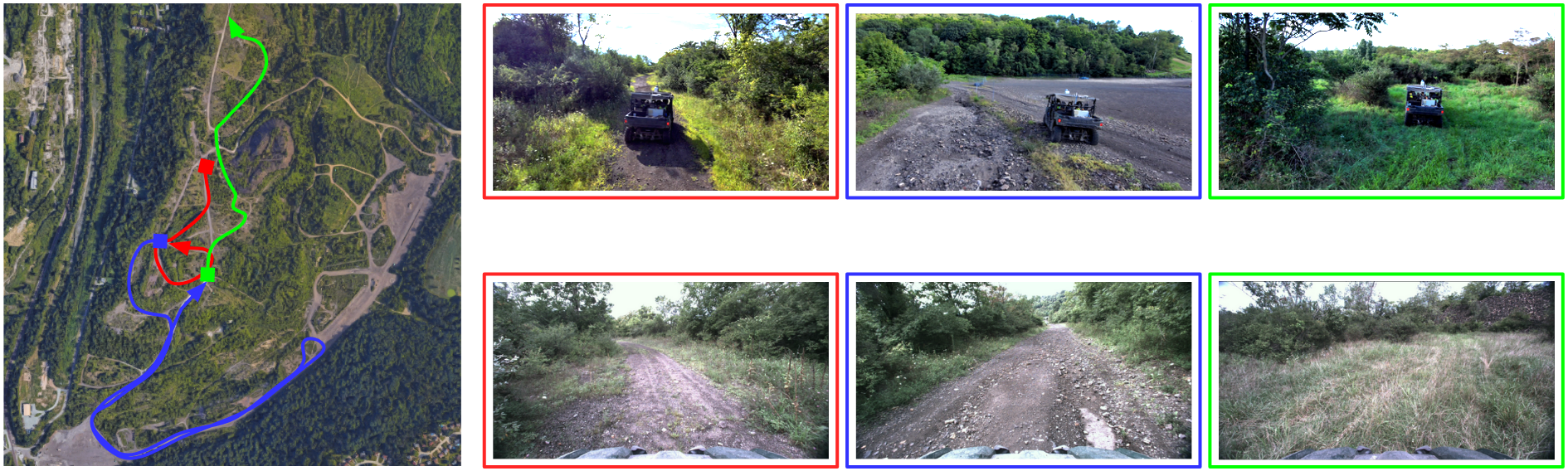}
    \caption{Overview of large-scale navigation experiments. The leftmost figure shows a satellite view of the three courses: red, blue, and green. The top row shows sample aerial views of the courses, while the bottom row shows sample first-person robot images. The color of the rectangle around the picture denotes the course.}
    \vspace{-0.25in}
    \label{fig:large_nav_experiments}
\end{figure}

\textbf{Do learned costmaps improve \textit{short-scale} navigation?}
We deploy our learned costmaps within a full navigation stack in two different navigation tasks, and compare the performance against a navigation stack which uses just a baseline geometric occupancy-based costmap. When using our learned costmaps, we compose them with the occupancy-based costmap to provide basic obstacle avoidance capabilities. We set the lethal height to be 1.5 m and all unknown regions to have a cost of 0.5 for all experiments.

We set up two navigation courses for the ATV, ATV-Warehouse (Figure \ref{fig:nav_experiments}) and ATV-Turnpike, where the task is to simply move 400m and 200m straight ahead, respectively. We design these controlled courses such that a straight path would lead the robot to go over rougher terrain or patches of vegetation, but reasoning about the terrain would lead the robot to take a detour to stay on smoother trails. For ATV-Warehouse, we run five trials with our learned costmap and five trials with the baseline costmap, and for ATV-Turnpike, we run three trials for each costmap.

The baseline costmap leads the robot to navigate in a straight line (with some noise), which leads straight into the patch of vegetation. Additionally, this baseline stack leads to less consistent navigation, which occurs because most of the terrain in front of the robot appears to be ``traversable," since there are no obstacles above the 1.5 m lethal height, which leads to many different ``optimal" paths. On the other hand, our learned costmaps cause the robot to consistently navigate around the patch, staying on marked paths, and even moving away from parts of the path with gravel towards smoother sections of the trail. 

We measure the cross-track error (CTE) as a way of measuring the deviation of the chosen path from a na\"ive, straight path to the goal. In this case, lower is \textit{not} necessarily better, since the experiment is explicitly designed for the robot to deviate from the straight path to find the smoother trails if it is able to reason about different terrains. We show the preferred paths in Figure \ref{fig:nav_experiments}, and numerical results in Table \ref{tab:nav_trials}. 
% Details of the ATV-Turnpike experiment are available in the Appendix. 
ATV-Turnpike experiment details are in the Appendix.

\begin{table}
    \vspace{0.05in}
    \centering
    \begin{tabular}{|c|c|c|c|}
    \hline
        Course & Nav. Stack & Avg. CTE (m)\\ \hline \hline
        ATV-Warehouse & Baseline & $1.29 \pm 1.02$\\ \hline
        ATV-Warehouse & HDIF (Ours) & $3.39 \pm 0.94$\\ \hline
        ATV-Pole & Baseline & $2.44 \pm 2.25$\\ \hline
        ATV-Pole & HDIF (Ours) & $2.68 \pm 0.41$\\ \hline
    \end{tabular}
    \caption{Avg. Cross-track error (CTE) between planned paths and the shortest path to goal. Learned costmaps resulted in trajectories that deviated more from the shortest path to avoid areas of higher cost, e.g. grass or gravel.}
    \label{tab:nav_trials}
    \vspace{-0.25in}
\end{table}

\textbf{Do learned costmaps improve large-scale navigation?}
To provide a direct measurement of navigation performance, we set up three large scale navigation experiments in three challenging off-road courses, with waypoints every 50 m, and measure our navigation performance using number of interventions, as is common practice \cite{krusi2017driving, bojarski2016end}. The safety driver was directed to only intervene when either: a) the robot missed a waypoint by over 4 m, or b) a collision with a non-traversable obstacle was imminent. These courses (Figure \ref{fig:large_nav_experiments}) include flat and hilly terrain, loose material ranging from fine gravel to large cobbles, and vegetation of various dimensions. For both navigation stacks, we set the target speed to 3.5 m/s, lethal height to 0.5 m and fill unknown values of our costmap to 0.5. We observe that in all three courses, our navigation stack with learned costmaps outperforms the baseline stack in terms of number of interventions, as detailed in Table \ref{tab:large_nav_trials}. Our learned costmaps lead the robot to stay on smoother paths and avoid vegetation if possible, which leads to an overall decrease in intervention events. In practice, this leads to the robot taking wider turns, staying on clear paths, and avoiding rough features that lead to bumpy navigation, such as large cobbles.  

\begin{table}
    \vspace{0.05in}
    \centering
    \begin{tabular}{|c|c|c|c|c|}
    \hline
        Navigation Stack & Course & Interventions & Course Length (m) \\ \hline \hline
        Baseline & Red & 7 & 400 \\ \hline
        HDIF (Ours) & Red & \textbf{3} & 400 \\ \hline \hline
        Baseline & Blue & 9 & 3150 \\ \hline
        HDIF (Ours) & Blue & \textbf{6} & 3150\\ \hline \hline
        Baseline & Green & 11 & 950\\ \hline
        HDIF (Ours) & Green & \textbf{7} & 950\\ \hline
    \end{tabular}
    \caption{Number of interventions (lower is better) in three large-scale navigation courses with varying degrees of difficulty and different types of environment features, such as slopes, vegetation, and tight turns. Our How Does It Feel (HDIF) navigation stack achieves lower number of interventions in all courses compared to the baseline stack.}
    \label{tab:large_nav_trials}
    \vspace{-0.25in}
\end{table}

\textbf{How well does our method transfer to other robots?}

We set up two additional short-scale navigation experiments using the Warthog, with a fine-tuned model as detailed in Section \ref{sec:data}. We used a separate geometry-based autonomy stack \cite{arl_stack} as the baseline, and for our stack we simply composed their lidar-based binary costmap with our learned costmap. Despite the Warthog being a skid-steer robot, we treat it as if it had an Ackermann steering geometry, similar to the ATV, since its cameras only face forward. 

In the first experiment, the goal is to drive 50 m to a goal located straight ahead, where the straight path goes through grass, but there is a smooth concrete path available surrounding this patch of vegetation. Similar to what we observe in the ATV experiments, our costmaps allow the Warthog to reason about the traversability properties of the smooth concrete and the vegetation, and the robot chooses to take the smooth path around the grass to reach the goal. 

In the second experiment, we artificially create two different paths at a fork in a forest trail. We litter the path on the left with small logs, fallen leaves, and small rocks, while we clear the path on the right of all obstacles. We observe that with the baseline stack, the robot uniformly chooses either of the paths to get to a goal at the other end of the fork in the trail. On the other hand, with our costmaps, the robot consistently chooses the clear path on the right to get to the other side of the trail. Note that a visual semantic classifier would label both of these trails as belonging to the same class, but the subtle differences in the features of the trails cause different amounts of roughness. 

In both experiments, it is clear that our costmaps lead the robot to choose smoother paths by reasoning about the different traversability properties of different terrains. We urge the reader to visit our website for experiment videos, and our Appendix for more experiment details.

%===============================================================================

%===============================================================================

\section{Conclusion and Future Work}
\label{sec:conclusion}

    In this work we present a costmap prediction system that predicts what the interactions between the ground and the robot feel like based on environmental characteristics and robot dynamics. To achieve this, we train a network that combines exteroception and information about the robot's velocity to predict a traversability cost derived directly from robot proprioception. We demonstrate that our costmaps enable for more nuanced and diverse navigation behaviors compared to a common baseline. Future work includes online adaptation for better generalization, uncertainty estimation, as well as improved representations for exteroceptive data to overcome our current perception limitations. 

\addtolength{\textheight}{-1cm}   % This command serves to balance the column lengths

%%%%%%%%%%%%%%%%%%%%%%%%%%%%%%%%%%%%%%%%%%%%%%%%%%%%%%%%%%%%%%%%%%%%%%%%%%%%%%%%

%%%%%%%%%%%%%%%%%%%%%%%%%%%%%%%%%%%%%%%%%%%%%%%%%%%%%%%%%%%%%%%%%%%%%%%%%%%%%%%%



\section*{ACKNOWLEDGMENT}

We thank Matthew Sivaprakasam, Sean Wang, Eli Lancaster, Andrew Saba, and Ananya Rao for their help during field testing, and the members of AirLab for insightful discussions.

%%%%%%%%%%%%%%%%%%%%%%%%%%%%%%%%%%%%%%%%%%%%%%%%%%%%%%%%%%%%%%%%%%%%%%%%%%%%%%%%

% References are important to the reader; therefore, each citation must be complete and correct. If at all possible, references should be commonly available publications.

{
\bibliographystyle{IEEEtran}
\bibliography{references}

\begin{thebibliography}{10}
\providecommand{\url}[1]{#1}
\csname url@rmstyle\endcsname
\providecommand{\newblock}{\relax}
\providecommand{\bibinfo}[2]{#2}
\providecommand\BIBentrySTDinterwordspacing{\spaceskip=0pt\relax}
\providecommand\BIBentryALTinterwordstretchfactor{4}
\providecommand\BIBentryALTinterwordspacing{\spaceskip=\fontdimen2\font plus
\BIBentryALTinterwordstretchfactor\fontdimen3\font minus
  \fontdimen4\font\relax}
\providecommand\BIBforeignlanguage[2]{{%
\expandafter\ifx\csname l@#1\endcsname\relax
\typeout{** WARNING: IEEEtran.bst: No hyphenation pattern has been}%
\typeout{** loaded for the language `#1'. Using the pattern for}%
\typeout{** the default language instead.}%
\else
\language=\csname l@#1\endcsname
\fi
#2}}

\bibitem{fu2021coupling}
Z.~Fu, A.~Kumar, A.~Agarwal, H.~Qi, J.~Malik, and D.~Pathak, ``Coupling vision
  and proprioception for navigation of legged robots,'' \emph{arXiv preprint
  arXiv:2112.02094}, 2021.

\bibitem{fan2021step}
D.~D. Fan, K.~Otsu, Y.~Kubo, A.~Dixit, J.~Burdick, and A.-A. Agha-Mohammadi,
  ``Step: Stochastic traversability evaluation and planning for risk-aware
  off-road navigation,'' \emph{arXiv preprint arXiv:2103.02828}, 2021.

\bibitem{maturana2018real}
D.~Maturana, P.-W. Chou, M.~Uenoyama, and S.~Scherer, ``Real-time semantic
  mapping for autonomous off-road navigation,'' in \emph{Field and Service
  Robotics}.\hskip 1em plus 0.5em minus 0.4em\relax Springer, 2018, pp.
  335--350.

\bibitem{kim2018season}
D.-K. Kim, D.~Maturana, M.~Uenoyama, and S.~Scherer, ``Season-invariant
  semantic segmentation with a deep multimodal network,'' in \emph{Field and
  service robotics}.\hskip 1em plus 0.5em minus 0.4em\relax Springer, 2018, pp.
  255--270.

\bibitem{lalonde2006natural}
J.-F. Lalonde, N.~Vandapel, D.~F. Huber, and M.~Hebert, ``Natural terrain
  classification using three-dimensional ladar data for ground robot
  mobility,'' \emph{Journal of field robotics}, vol.~23, no.~10, pp. 839--861,
  2006.

\bibitem{krusi2017driving}
P.~Kr{\"u}si, P.~Furgale, M.~Bosse, and R.~Siegwart, ``Driving on point clouds:
  Motion planning, trajectory optimization, and terrain assessment in generic
  nonplanar environments,'' \emph{Journal of Field Robotics}, vol.~34, no.~5,
  pp. 940--984, 2017.

\bibitem{gennery1999traversability}
D.~B. Gennery, ``Traversability analysis and path planning for a planetary
  rover,'' \emph{Autonomous Robots}, vol.~6, no.~2, pp. 131--146, 1999.

\bibitem{kahn2021badgr}
G.~Kahn, P.~Abbeel, and S.~Levine, ``Badgr: An autonomous self-supervised
  learning-based navigation system,'' \emph{IEEE Robotics and Automation
  Letters}, vol.~6, no.~2, pp. 1312--1319, 2021.

\bibitem{wellhausen2019should}
L.~Wellhausen, A.~Dosovitskiy, R.~Ranftl, K.~Walas, C.~Cadena, and M.~Hutter,
  ``Where should i walk? predicting terrain properties from images via
  self-supervised learning,'' \emph{IEEE Robotics and Automation Letters},
  vol.~4, no.~2, pp. 1509--1516, 2019.

\bibitem{mildenhall2020nerf}
B.~Mildenhall, P.~P. Srinivasan, M.~Tancik, J.~T. Barron, R.~Ramamoorthi, and
  R.~Ng, ``Nerf: Representing scenes as neural radiance fields for view
  synthesis,'' in \emph{European conference on computer vision}.\hskip 1em plus
  0.5em minus 0.4em\relax Springer, 2020, pp. 405--421.

\bibitem{tancik2020fourier}
M.~Tancik, P.~Srinivasan, B.~Mildenhall, S.~Fridovich-Keil, N.~Raghavan,
  U.~Singhal, R.~Ramamoorthi, J.~Barron, and R.~Ng, ``Fourier features let
  networks learn high frequency functions in low dimensional domains,''
  \emph{Advances in Neural Information Processing Systems}, vol.~33, pp.
  7537--7547, 2020.

\bibitem{howard2006towards}
A.~Howard, M.~Turmon, L.~Matthies, B.~Tang, A.~Angelova, and E.~Mjolsness,
  ``Towards learned traversability for robot navigation: From underfoot to the
  far field,'' \emph{Journal of Field Robotics}, vol.~23, no. 11-12, pp.
  1005--1017, 2006.

\bibitem{angelova2007learning}
A.~Angelova, L.~Matthies, D.~Helmick, and P.~Perona, ``Learning and prediction
  of slip from visual information,'' \emph{Journal of Field Robotics}, vol.~24,
  no.~3, pp. 205--231, 2007.

\bibitem{hadsell2009learning}
R.~Hadsell, P.~Sermanet, J.~Ben, A.~Erkan, M.~Scoffier, K.~Kavukcuoglu,
  U.~Muller, and Y.~LeCun, ``Learning long-range vision for autonomous off-road
  driving,'' \emph{Journal of Field Robotics}, vol.~26, no.~2, pp. 120--144,
  2009.

\bibitem{dupont2008frequency}
E.~M. Dupont, C.~A. Moore, E.~G. Collins, and E.~Coyle, ``Frequency response
  method for terrain classification in autonomous ground vehicles,''
  \emph{Autonomous Robots}, vol.~24, no.~4, pp. 337--347, 2008.

\bibitem{konolige2009mapping}
K.~Konolige, M.~Agrawal, M.~R. Blas, R.~C. Bolles, B.~Gerkey, J.~Sola, and
  A.~Sundaresan, ``Mapping, navigation, and learning for off-road traversal,''
  \emph{Journal of Field Robotics}, vol.~26, no.~1, pp. 88--113, 2009.

\bibitem{bajracharya2009autonomous}
M.~Bajracharya, A.~Howard, L.~H. Matthies, B.~Tang, and M.~Turmon, ``Autonomous
  off-road navigation with end-to-end learning for the {LAGR} program,''
  \emph{Journal of Field Robotics}, vol.~26, no.~1, pp. 3--25, 2009.

\bibitem{stavens2006self}
D.~Stavens and S.~Thrun, ``A self-supervised terrain roughness estimator for
  off-road autonomous driving,'' in \emph{Proceedings of the Twenty-Second
  Conference on Uncertainty in Artificial Intelligence}, 2006, pp. 469--476.

\bibitem{jackel2006darpa}
L.~D. Jackel, E.~Krotkov, M.~Perschbacher, J.~Pippine, and C.~Sullivan, ``The
  darpa lagr program: Goals, challenges, methodology, and phase i results,''
  \emph{Journal of Field robotics}, vol.~23, no. 11-12, pp. 945--973, 2006.

\bibitem{thrun2006stanley}
S.~Thrun, M.~Montemerlo, H.~Dahlkamp, D.~Stavens, A.~Aron, J.~Diebel, P.~Fong,
  J.~Gale, M.~Halpenny, G.~Hoffmann, \emph{et~al.}, ``Stanley: The robot that
  won the darpa grand challenge,'' \emph{Journal of field Robotics}, vol.~23,
  no.~9, pp. 661--692, 2006.

\bibitem{long2015fully}
J.~Long, E.~Shelhamer, and T.~Darrell, ``Fully convolutional networks for
  semantic segmentation,'' in \emph{Proceedings of the IEEE conference on
  computer vision and pattern recognition}, 2015, pp. 3431--3440.

\bibitem{shaban2022semantic}
A.~Shaban, X.~Meng, J.~Lee, B.~Boots, and D.~Fox, ``Semantic terrain
  classification for off-road autonomous driving,'' in \emph{Conference on
  Robot Learning}.\hskip 1em plus 0.5em minus 0.4em\relax PMLR, 2022, pp.
  619--629.

\bibitem{shah2022rapid}
D.~Shah, B.~Eysenbach, N.~Rhinehart, and S.~Levine, ``Rapid exploration for
  open-world navigation with latent goal models,'' in \emph{Conference on Robot
  Learning}.\hskip 1em plus 0.5em minus 0.4em\relax PMLR, 2022, pp. 674--684.

\bibitem{williams2017information}
G.~Williams, N.~Wagener, B.~Goldfain, P.~Drews, J.~M. Rehg, B.~Boots, and E.~A.
  Theodorou, ``Information theoretic mpc for model-based reinforcement
  learning,'' in \emph{2017 IEEE International Conference on Robotics and
  Automation (ICRA)}.\hskip 1em plus 0.5em minus 0.4em\relax IEEE, 2017, pp.
  1714--1721.

\bibitem{triest2022tartandrive}
S.~Triest, M.~Sivaprakasam, S.~J. Wang, W.~Wang, A.~M. Johnson, and S.~Scherer,
  ``Tartandrive: A large-scale dataset for learning off-road dynamics models,''
  in \emph{2022 International Conference on Robotics and Automation (ICRA)},
  2022, pp. 2546--2552.

\bibitem{sivaprakasam2021improving}
M.~Sivaprakasam, S.~Triest, W.~Wang, P.~Yin, and S.~Scherer, ``Improving
  off-road planning techniques with learned costs from physical interactions,''
  in \emph{2021 IEEE International Conference on Robotics and Automation
  (ICRA)}.\hskip 1em plus 0.5em minus 0.4em\relax IEEE, 2021, pp. 4844--4850.

\bibitem{fan2021learning}
D.~D. Fan, A.-A. Agha-Mohammadi, and E.~A. Theodorou, ``Learning risk-aware
  costmaps for traversability in challenging environments,'' \emph{IEEE
  Robotics and Automation Letters}, vol.~7, no.~1, pp. 279--286, 2021.

\bibitem{cai2022risk}
X.~Cai, M.~Everett, J.~Fink, and J.~P. How, ``Risk-aware off-road navigation
  via a learned speed distribution map,'' \emph{arXiv preprint
  arXiv:2203.13429}, 2022.

\bibitem{triest2023learning}
S.~Triest, M.~G. Castro, P.~Maheshwari, M.~Sivaprakasam, W.~Wang, and
  S.~Scherer, ``Learning risk-aware costmaps via inverse reinforcement learning
  for off-road navigation,'' \emph{arXiv preprint arXiv:2302.00134}, 2023.

\bibitem{sathyamoorthy2022terrapn}
A.~J. Sathyamoorthy, K.~Weerakoon, T.~Guan, J.~Liang, and D.~Manocha,
  ``Terrapn: Unstructured terrain navigation using online self-supervised
  learning,'' in \emph{2022 IEEE/RSJ International Conference on Intelligent
  Robots and Systems (IROS)}.\hskip 1em plus 0.5em minus 0.4em\relax IEEE,
  2022, pp. 7197--7204.

\bibitem{yao2022rca}
X.~Yao, J.~Zhang, and J.~Oh, ``Rca: Ride comfort-aware visual navigation via
  self-supervised learning,'' in \emph{2022 IEEE/RSJ International Conference
  on Intelligent Robots and Systems (IROS)}.\hskip 1em plus 0.5em minus
  0.4em\relax IEEE, 2022, pp. 7847--7852.

\bibitem{waibel2022rough}
G.~G. Waibel, T.~L{\"o}w, M.~Nass, D.~Howard, T.~Bandyopadhyay, and P.~V.~K.
  Borges, ``How rough is the path? terrain traversability estimation for local
  and global path planning,'' \emph{IEEE Transactions on Intelligent
  Transportation Systems}, vol.~23, no.~9, pp. 16\,462--16\,473, 2022.

\bibitem{welch1967use}
P.~Welch, ``The use of fast fourier transform for the estimation of power
  spectra: a method based on time averaging over short, modified
  periodograms,'' \emph{IEEE Transactions on audio and electroacoustics},
  vol.~15, no.~2, pp. 70--73, 1967.

\bibitem{chang2018pyramid}
J.-R. Chang and Y.-S. Chen, ``Pyramid stereo matching network,'' in
  \emph{Proceedings of the IEEE conference on computer vision and pattern
  recognition}, 2018, pp. 5410--5418.

\bibitem{wang2020tartanvo}
W.~Wang, Y.~Hu, and S.~Scherer, ``Tartanvo: A generalizable learning-based
  vo,'' \emph{arXiv preprint arXiv:2011.00359}, 2020.

\bibitem{rahaman2019spectral}
N.~Rahaman, A.~Baratin, D.~Arpit, F.~Draxler, M.~Lin, F.~Hamprecht, Y.~Bengio,
  and A.~Courville, ``On the spectral bias of neural networks,'' in
  \emph{International Conference on Machine Learning}.\hskip 1em plus 0.5em
  minus 0.4em\relax PMLR, 2019, pp. 5301--5310.

\bibitem{he2016deep}
K.~He, X.~Zhang, S.~Ren, and J.~Sun, ``Deep residual learning for image
  recognition,'' in \emph{Proceedings of the IEEE conference on computer vision
  and pattern recognition}, 2016, pp. 770--778.

\bibitem{kingma2014adam}
D.~P. Kingma and J.~Ba, ``Adam: A method for stochastic optimization,''
  \emph{arXiv preprint arXiv:1412.6980}, 2014.

\bibitem{zhao2021super}
S.~Zhao, H.~Zhang, P.~Wang, L.~Nogueira, and S.~Scherer, ``Super odometry:
  Imu-centric lidar-visual-inertial estimator for challenging environments,''
  in \emph{2021 IEEE/RSJ International Conference on Intelligent Robots and
  Systems (IROS)}.\hskip 1em plus 0.5em minus 0.4em\relax IEEE, 2021, pp.
  8729--8736.

\bibitem{gregory2016application}
J.~Gregory, J.~Fink, E.~Stump, J.~Twigg, J.~Rogers, D.~Baran, N.~Fung, and
  S.~Young, ``Application of multi-robot systems to disaster-relief scenarios
  with limited communication,'' in \emph{Field and Service Robotics}.\hskip 1em
  plus 0.5em minus 0.4em\relax Springer, 2016, pp. 639--653.

\bibitem{clearpath_robotics_2021}
\BIBentryALTinterwordspacing
``Warthog unmanned ground vehicle robot - clearpath,'' Jan 2021. [Online].
  Available:
  \url{https://clearpathrobotics.com/warthog-unmanned-ground-vehicle-robot/}
\BIBentrySTDinterwordspacing

\bibitem{fankhauser2018robust}
P.~Fankhauser, M.~Bjelonic, C.~D. Bellicoso, T.~Miki, and M.~Hutter, ``Robust
  rough-terrain locomotion with a quadrupedal robot,'' in \emph{2018 IEEE
  International Conference on Robotics and Automation (ICRA)}.\hskip 1em plus
  0.5em minus 0.4em\relax IEEE, 2018, pp. 5761--5768.

\bibitem{bojarski2016end}
M.~Bojarski, D.~Del~Testa, D.~Dworakowski, B.~Firner, B.~Flepp, P.~Goyal, L.~D.
  Jackel, M.~Monfort, U.~Muller, J.~Zhang, \emph{et~al.}, ``End to end learning
  for self-driving cars,'' \emph{arXiv preprint arXiv:1604.07316}, 2016.

\bibitem{arl_stack}
\BIBentryALTinterwordspacing
``Arl autonomy stack,'' 2022, accessed 13-September-2022. [Online]. Available:
  \url{https://www.arl.army.mil/business/collaborative-alliances/current-cras/sara-cra/sara-overview/}
\BIBentrySTDinterwordspacing

\bibitem{Mai-2020}
J.~Mai, ``System design, modelling, and control for an off-road autonomous
  ground vehicle,'' Master's thesis, Carnegie Mellon University, Pittsburgh,
  PA, July 2020.

\bibitem{marcel2010torchvision}
S.~Marcel and Y.~Rodriguez, ``Torchvision the machine-vision package of
  torch,'' in \emph{Proceedings of the 18th ACM international conference on
  Multimedia}, 2010, pp. 1485--1488.

\end{thebibliography}
}

\appendix

\subsection{Traversability Cost Analysis}
\label{sec:traversability_appendix}
	
\begin{figure}[b]
        \centering
        \includegraphics[width=0.5\linewidth]{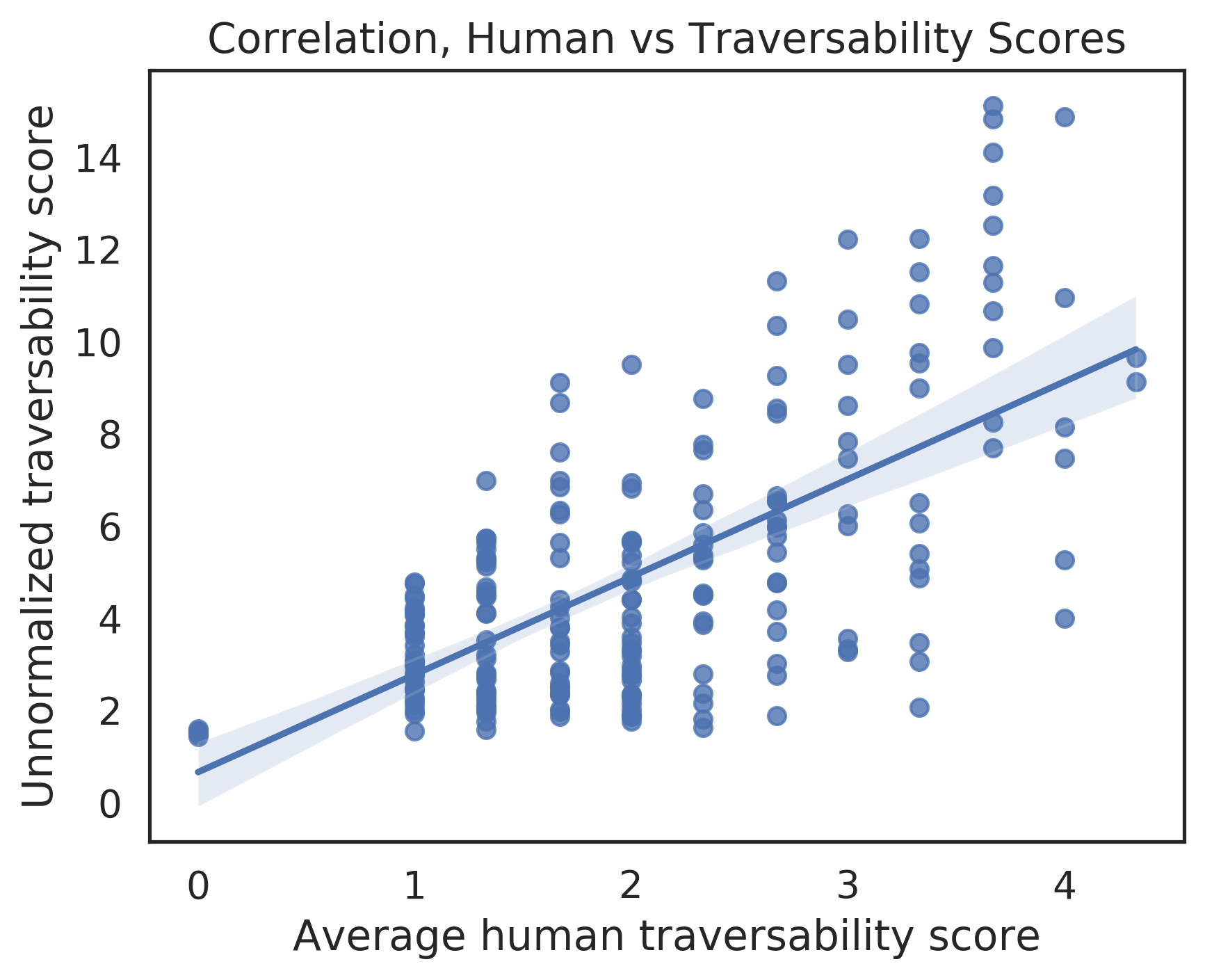}
        \caption{Correlation between the average of three human-labeled traversability scores (1-5) and our IMU traversability cost function. There is a strong correlation between the two scores, with a Pearson correlation coefficient of 0.66.}
        \label{fig:traversability_cost}
    \end{figure}

	To decide the best frequency band for our traversability cost, we collect an evaluation set of 220 5-second trajectories on our robot, spanning smooth, bumpy, sloped, and grassy trails at different speeds. Each of these trajectories is annotated by three human labelers with a score of 1-5 as to how traversable they were, with 5 meaning most difficult to traverse. We calculate the correlation between the average human score and different frequency ranges to tune the frequency range used in our cost function. We found the best frequency range to be 1-30 Hz, which had a Pearson correlation coefficient of 0.66, as shown in Figure \ref{fig:traversability_cost}. Finally, we use the evaluation set to obtain statistics that we use to normalize the traversability cost function between 0 and 1.

\subsection{Mapping}
\label{sec:mapping_appendix}
    
    Our method relies on having a dense, colorized point cloud from which we can extract corresponding color and height information about the environment. We experiment with two different setups. When the robot contains a lidar which produces dense point clouds, we can simply colorize these points using the RGB information from a calibrated monocular camera. When the robot is not outfitted with a lidar, or the lidar produces only sparse point clouds in the near-range, we use a stereo matching network \cite{chang2018pyramid} to obtain a disparirty image, from which we estimate the camera odometry using TartanVO \cite{wang2020tartanvo}, a learning-based visual odometry algorithm. We use this odometry and the RGB data to register and colorize a dense point cloud which we then project into a top-down view.
    
    In our setup, we use a $12 \times 12$ meter local map with a resolution of $0.02$ meters. This results in a $600 \times 600$ cell local map consisting of eight channels: three channels for RGB information, four channels containing the minimum, maximum, mean, and standard deviation of the height of the points in each cell, and a channel describing whether each cell in the local map is unknown. The dimensions of the local map match the dimensions of the learned costmap that will be used directly for path planning. 
	
\subsection{Robot Platform}
\label{sec:robot_appendix}

    We perform experiments on two different ground robots: a large, Yamaha Viking side-by-side all-terrain vehicle (ATV) modified for autonomous driving by Mai et al. \cite{Mai-2020}, and a Clearpath Robotics Warthog unmanned ground vehicle (UGV). The ATV contains a front-facing Carnegie Robotics Multisense S21 stereo camera, a Velodyne Ultra Puck lidar, a NovAtel PROPAK-V3-RT2i GNSS unit providing IMU data and global pose estimates, as well as an onboard computer with an NVIDIA Geforce RTX3080 Laptop GPU. The Warthog UGV has two FLIR Blackfly S cameras providing a stereo pair with an approximately 53cm baseline, an Ouster OS1-64 LiDAR, a Microstrain 3DM-GX5-35 IMU which provides linear acceleration data and robot odometry measurements, and two Neousys Nuvo 7166GC onboard computers, each with an Intel i9 CPU and NVidia Tesla T4 GPU.
	
\subsection{Navigation Stack}
\label{sec:navigation_appendix}

    The kinematic bicycle model for the ATV vehicle and the Warthog UGV is shown in Algorithm \ref{algo:kbm}.
    % , while the skid-steer model for the Warthog UGV is shown in Algorithm \ref{algo:skid_steer}. 

    \begin{algorithm}
        \DontPrintSemicolon
        \caption{Kinematic Bicycle Model}
        \KwIn{Current state $s = [x, y, \theta, v, \delta$] (position, orientation, velocity and steering angle), Control $a = [v_{des}, \delta_{des}]$ (velocity and steering setpoints), Hyperparameters $L$ (wheelbase), $K_v$ (velocity gain), $K_\delta$ (steer angle gain)}
        \KwOut{Time derivative of state $\dot{s} = f(s, a)$}
        
        \begin{equation}
        \dot{s} = 
        \begin{bmatrix}
        \dot{x} \\
        \dot{y} \\
        \dot{\theta} \\
        \dot{v} \\
        \dot{\delta} \\
        \end{bmatrix}
        =\begin{bmatrix}
        v cos(\theta) \\
        v sin(\theta) \\
        \frac{v tan(\delta)}{L} \\
        K_v (v_{des} - v) \\
        K_\delta (\delta_{des} - \delta)
        \end{bmatrix}
        \end{equation} \;
        \label{algo:kbm}
    \end{algorithm}
    
    % \begin{algorithm}
    %     \DontPrintSemicolon
    %     \caption{Skid-Steer Model}
    %     \KwIn{Current state $s = [x, y, \theta, v, \omega]$ (position, orientation, linear and angular velocity), Control $a = [v_{des}, \omega_{des}]$ (linear and angular velocity setpoints), Hyperparameters $K_v$ (linear velocity gain), $K_\delta$ (angular velocity gain)}
    %     \KwOut{Time derivative of state $\dot{s} = f(s, a)$}
        
    %     \begin{equation}
    %     \dot{s} = 
    %     \begin{bmatrix}
    %     \dot{x} \\
    %     \dot{y} \\
    %     \dot{\theta} \\
    %     \dot{v} \\
    %     \dot{\omega} \\
    %     \end{bmatrix}
    %     =
    %     \begin{bmatrix}
    %     v cos(\theta) \\
    %     v sin(\theta) \\
    %     \omega \\
    %     K_v (v_{des} - v) \\
    %     K_\omega (\omega_{des} - \omega)
    %     \end{bmatrix}
    %     \end{equation} \;
    %     \label{algo:skid_steer}
    % \end{algorithm}

\label{sec:results_appendix}
\subsection{Training results}
We compared three models with different inputs: \texttt{patch}, \texttt{patch-vel}, and \texttt{patch-Fourier-vel}. All of these models used the Torchvision \cite{marcel2010torchvision} implementation of ResNet18 \cite{he2016deep}, and were trained using the Adam optimizer \cite{kingma2014adam}. The results are shown in Table \ref{tab:network_arch}. The training hyperparameters for all three models are specified in Table \ref{tab:hyperparameters}.

\begin{table}
    \centering
    \begin{tabular}{|c|c|c|}
    \hline
        Model & Train Loss ($\times10^{-2}$) & Val. Loss ($\times10^{-2}$)\\ \hline \hline
        Random & $22.0 \pm ~0.14$ & $22.0 \pm ~0.14$ \\ \hline
        Patch & $4.0 \pm ~0.059$ & $5.7 \pm ~0.43$ \\ \hline
        Patch-vel & $3.9 \pm ~0.074$ & $5.2 \pm ~0.15$ \\ \hline
        Patch-Fourier-vel & $\mathbf{3.8 \pm ~0.09}$ & $\mathbf{5.0 \pm ~0.11}$\\ \hline
    \end{tabular}
    \caption{Training and validation losses for three different models with different inputs. All models were trained with five random seeds. Adding velocity as an input improves training results, and the model which includes Fourier-parameterized velocity performs best (lower is better).}
    \label{tab:network_arch}
    % \vspace{-0.15in}
\end{table}

\begin{table}
    \centering
    \begin{tabular}{|c|c|c|} \hline %\toprule
        Hyperparameter & Value & Model \\ \hline \hline %\midrule
        Epochs & 50 & all\\ \hline %\midrule 
        Learning Rate & \num{3e-4} & all\\ \hline %\bottomrule
        $\gamma$ (Learning Rate Decay) & 0.99 & all\\ \hline
        MLP Num. Layers & 3  & all\\ \hline
        MLP Num. Units & 512  & all\\ \hline
        CNN Embedding Size & 512  & all\\ \hline
        $m$ (Number of Frequencies) & 16  & \texttt{patch-Fourier-vel}\\ \hline
        $\sigma$ (Frequency Scale) & 10  & \texttt{patch-Fourier-vel}\\ \hline
        
    \end{tabular}
    \caption{Hyperparameters used in training three models: \texttt{patch}, \texttt{patch-vel}, and \texttt{patch-Fourier-vel}.}
    \label{tab:hyperparameters}
\end{table}

We performed an ablation on the inputs to the network to verify whether adding height information improved performance over using just RGB data. As shown in Table \ref{tab:rgb-only}, adding height statistics results in significant improvement at training time, but similar performance at test time. We performed an additional ablation over the parameters for our Fourier frequency parameterization and found that  varying the scale of Fourier frequencies does not change performance significantly, and neither did varying the number of sampled frequencies $m$, as reported in Table \ref{tab:Fourier_param_ablation}.

\begin{table}[]
    \centering
    \begin{tabular}{|c|c|c|} \hline
    Model & Train Loss ($\times10^{-2}$) & Val. Loss ($\times10^{-2}$) \\ \hline \hline
    RGB-Fourier-vel   & $4.0 \pm ~0.083$ & $4.9 \pm ~0.06$ \\ \hline 
    Patch-Fourier-vel & $\mathbf{3.8 \pm ~0.09}$ & $5.0 \pm ~0.11$ \\ \hline
    \end{tabular}
    \caption{Ablation over model inputs. RGB-Fourier-vel uses only the RGB local map information, whereas Patch-Fourier-vel uses RGB and height statistics extracted from the point cloud. Using both RGB and height information results in slightly better performance in the training set, but comparable performance in the test set.}
    \label{tab:rgb-only}
\end{table}

\begin{table}
    \centering
    \begin{tabular}{|c|c|c|c|}
    \hline
        Hyperparameter & Value & Train Loss ($\times10^{-2}$) & Val. Loss ($\times10^{-2}$)\\ \hline \hline
        $\sigma$ & 1 & $3.8 \pm ~0.09$ & $4.9 \pm ~0.06$ \\ \hline
        $\sigma$ & 10 & $3.8 \pm ~0.09$ & $5.0 \pm ~0.11$ \\ \hline
        $\sigma$ & 100 & $3.7 \pm ~0.09$ & $5.2 \pm ~0.17$ \\ \hline \hline
        $m$ & 8 & $3.8 \pm ~0.04$ & $5.0 \pm ~0.24$ \\ \hline
        $m$ & 16 & $3.8 \pm ~0.09$ & $5.0 \pm ~0.11$ \\ \hline
        $m$ & 32 & $3.8 \pm ~0.06$ & $5.2 \pm ~0.37$ \\ \hline
    \end{tabular}
    \caption{Ablation over Fourier parameterization parameters. Each hyperparameter choice was evaluated for 5 random seeds. Changing the scale of either the sampled frequencies $\sigma$ or the number of frequencies samlpled $m$ did not significantly affect performance.}
    \label{tab:Fourier_param_ablation}
    \vspace{-0.15in}
\end{table}

\begin{figure}[ht]
\begin{subfigure}{.5\textwidth}
  \centering
  % include first image
  \includegraphics[width=.8\linewidth]{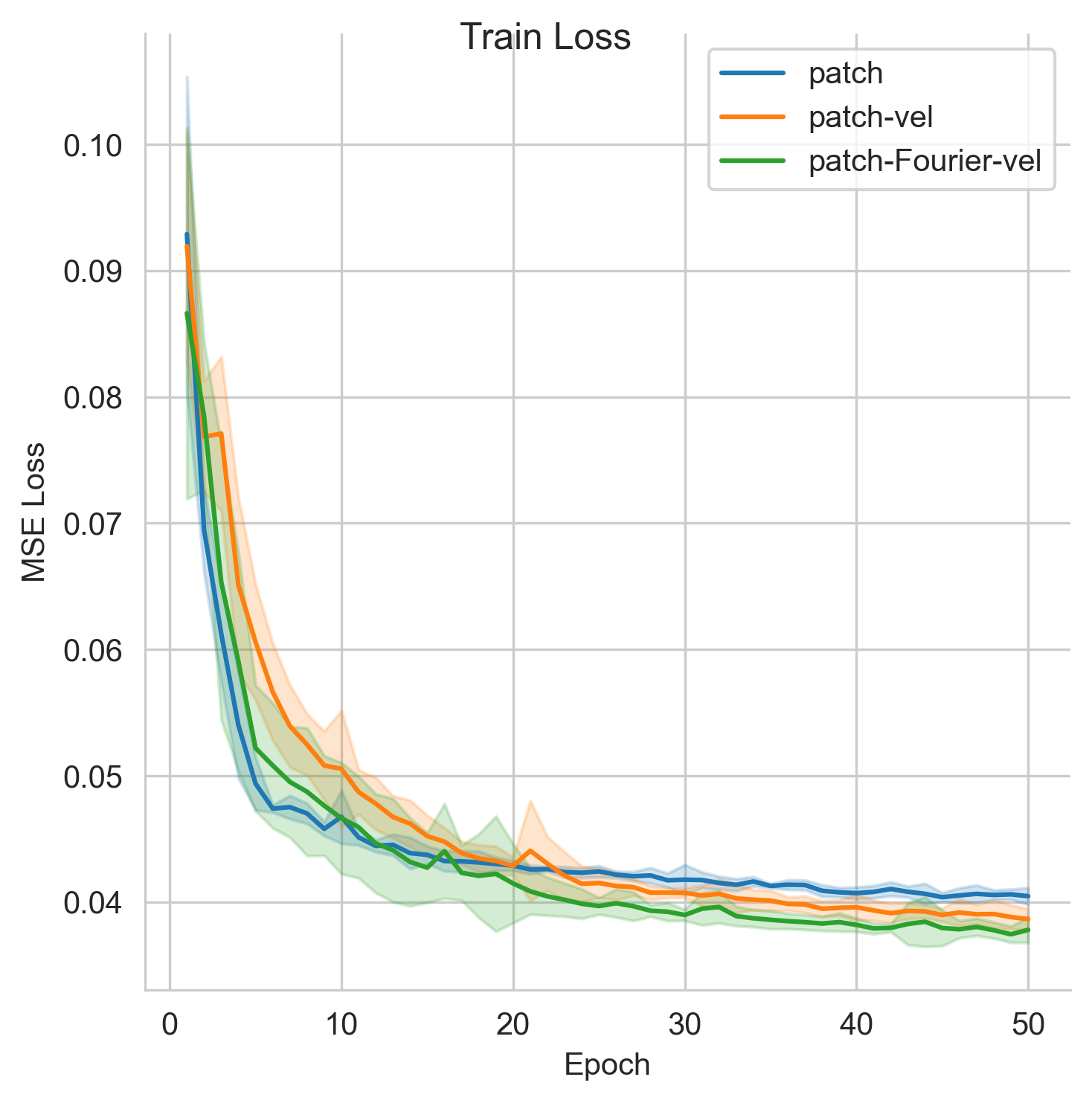}  
  \caption{Training loss curves.}
  \label{fig:train_loss}
\end{subfigure}
\begin{subfigure}{.5\textwidth}
  \centering
  % include second image
  \includegraphics[width=.8\linewidth]{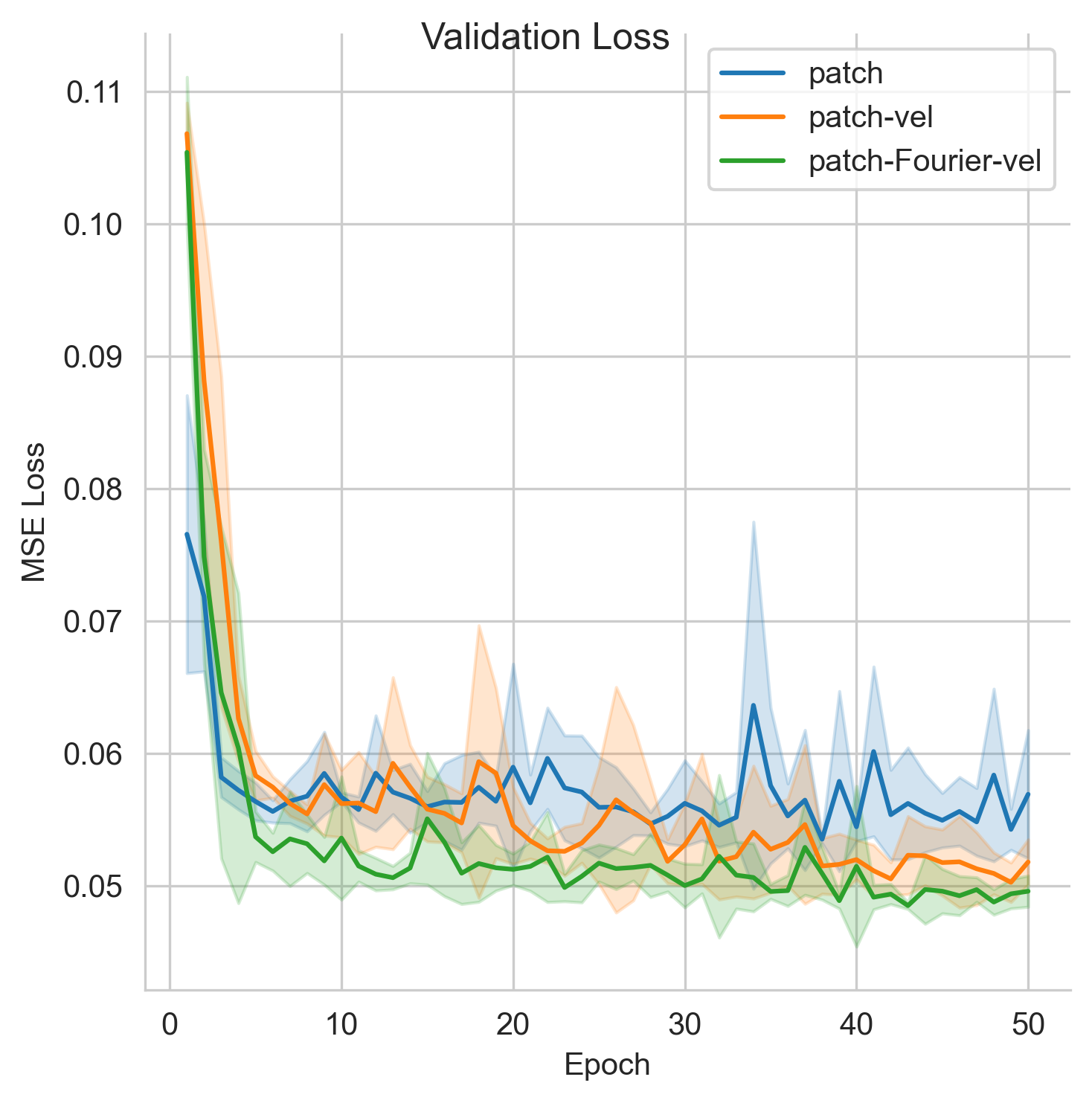}  
  \caption{Validation loss curves.}
  \label{fig:val_loss}
\end{subfigure}
\caption{Training and validation curves for the \texttt{patch}, \texttt{patch-vel}, and \texttt{patch-Fourier-vel} models. Including velocity as an input improves both training and validation loss, and including Fourier-parameterized velocity achieves the best results.}
\label{fig:models}
\end{figure}

\subsection{Velocity Experiments}
\begin{figure*}
    \centering
    % \vspace{-0.15in}
    \includegraphics[width=\linewidth]{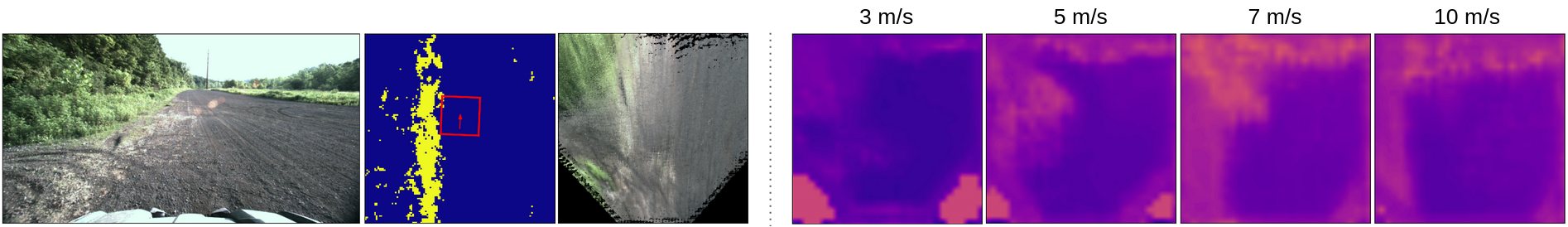}
    \caption{Costmaps at different velocities, as predicted by the \texttt{patch-Fourier-vel} model. The left block shows front-facing image, lethal-height costmap, and a top-down RGB map. The right block shows four costmaps of the same scenario at increasing speeds. Brighter means higher cost. Higher cost at the top of the costmaps is produced by artifacts in registration, as explained further in the Limitations section of this Appendix.}
    \label{fig:diff_vels}
\end{figure*}

We include costmaps predicted at different velocities in Figure \ref{fig:diff_vels}.

\subsection{Navigation Experiments}

\subsubsection{Short-Scale Navigation Experiments:}

\begin{figure}[ht]
\begin{subfigure}{.45\textwidth}
  \centering
  % include first image
  \includegraphics[width=.8\linewidth]{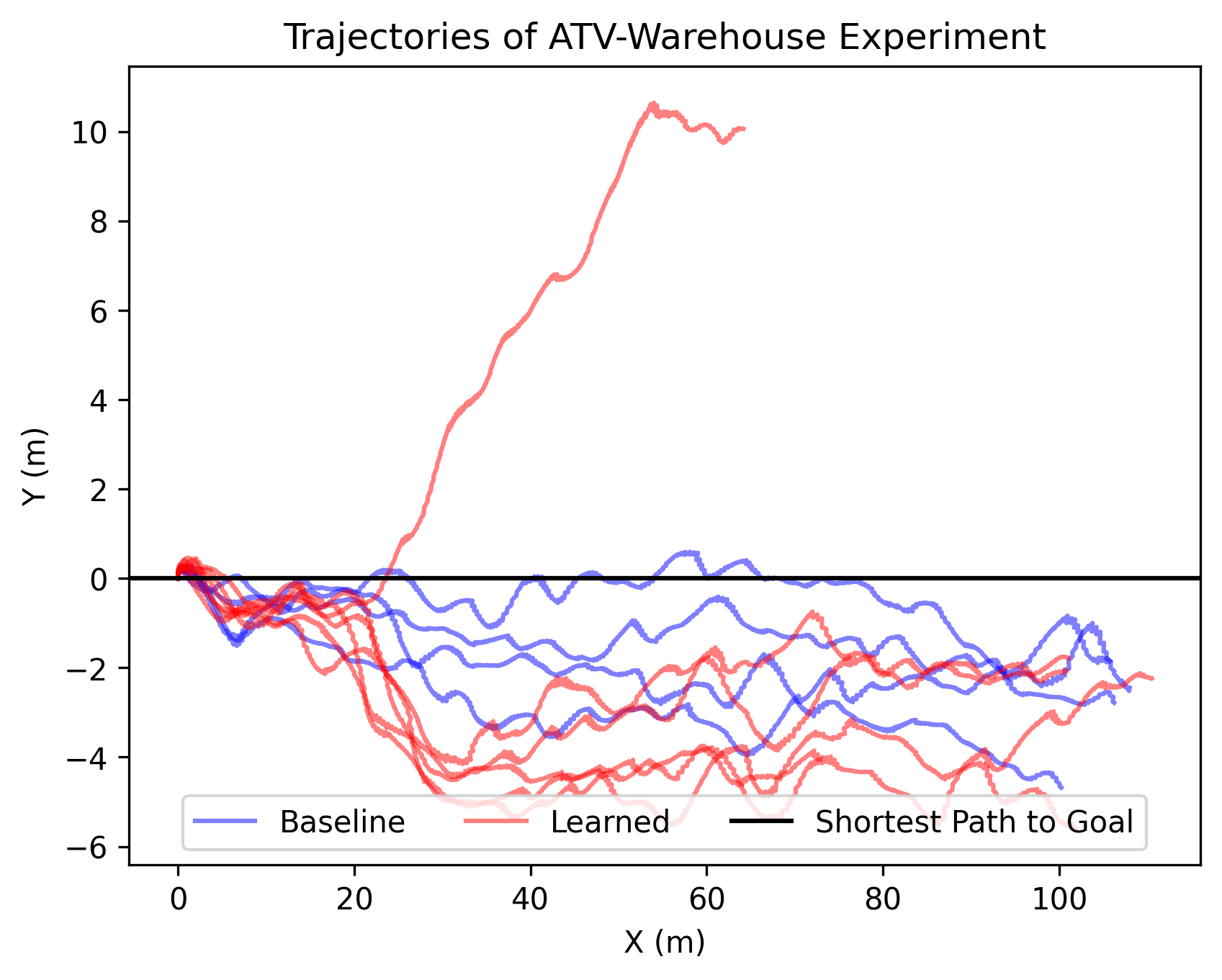}  
  \caption{Paths taken by the robots. Nominal trajectory (black) represents the straight path between the origin and the goal.}
  \label{fig:warehouose_paths}
\end{subfigure}
\hfill
\begin{subfigure}{.45\textwidth}
  \centering
  % include second image
  \includegraphics[width=.8\linewidth]{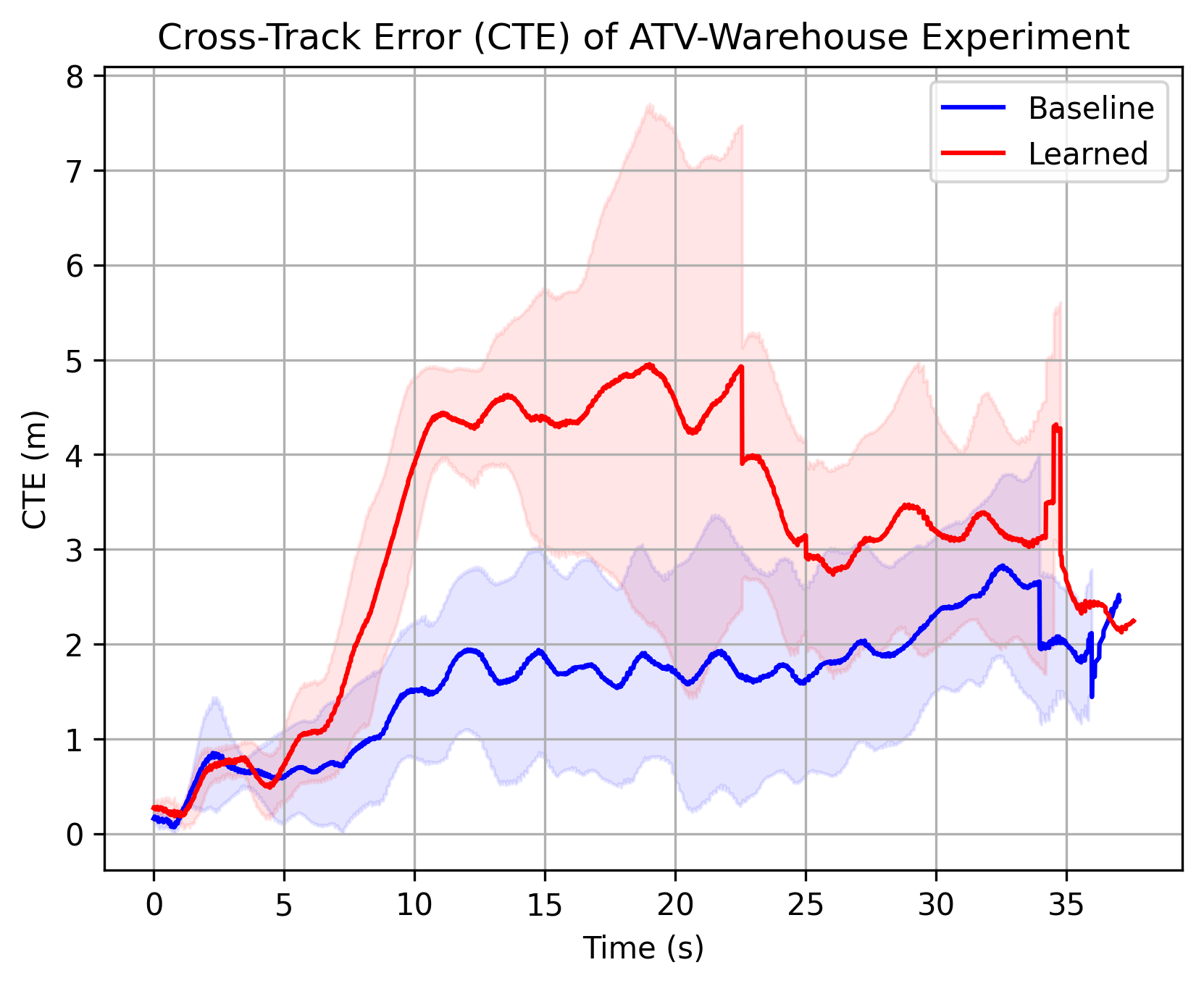}  
  \caption{Average cross-track error between the nominal trajectory and each of the paths taken by the robot over time.}
  \label{fig:warehouse_cte}
\end{subfigure}
\caption{ATV-Warehouse experiment.}
\label{fig:warehuose}
\end{figure}

\begin{figure}[ht]
\begin{subfigure}{.45\textwidth}
  \centering
  % include first image
  \includegraphics[width=.8\linewidth]{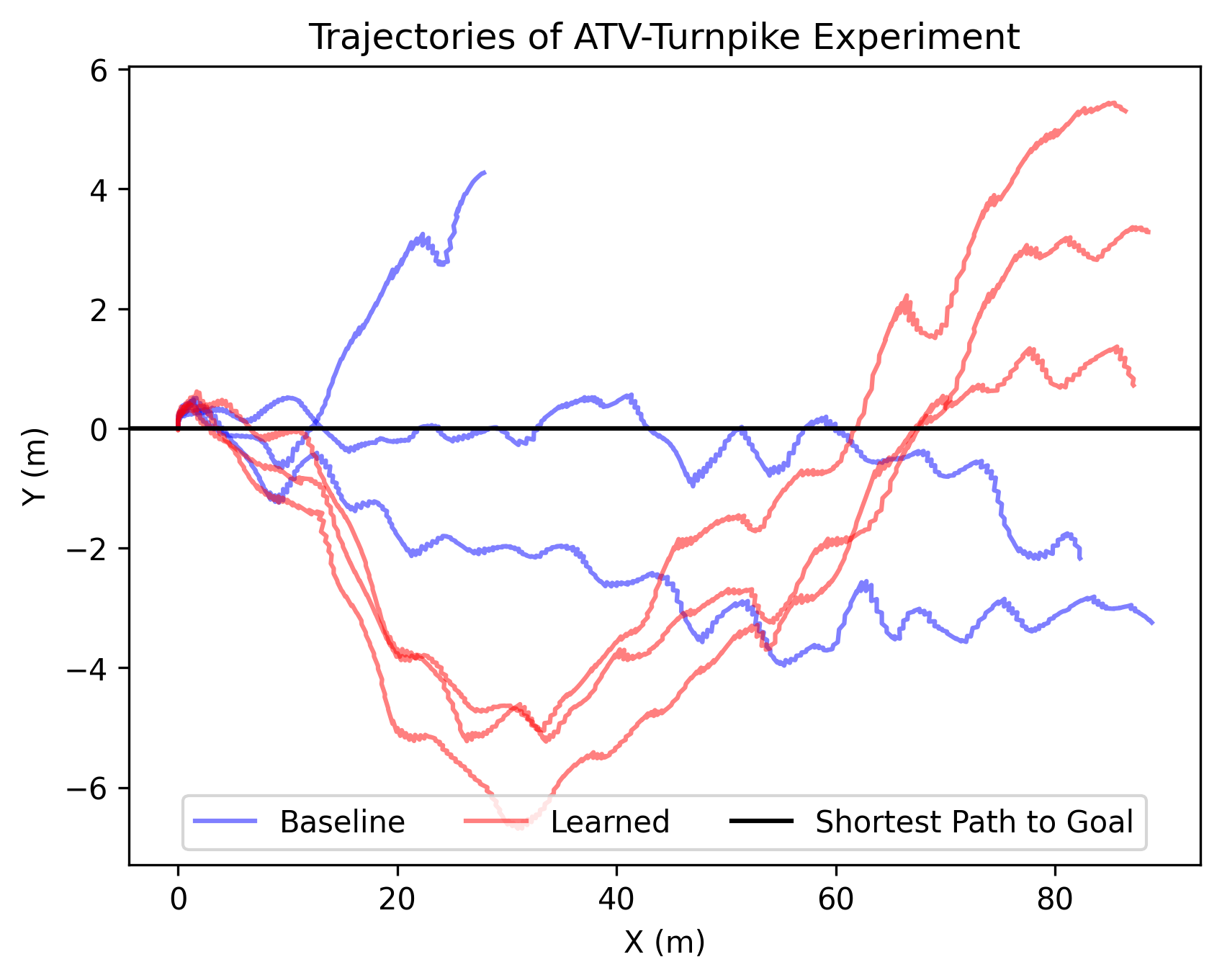}  
  \caption{Paths taken by the robots. Nominal trajectory (black) represents the straight path between the origin and the goal.}
  \label{fig:turnpike_paths}
\end{subfigure}
\hfill
\begin{subfigure}{.45\textwidth}
  \centering
  % include second image
  \includegraphics[width=.8\linewidth]{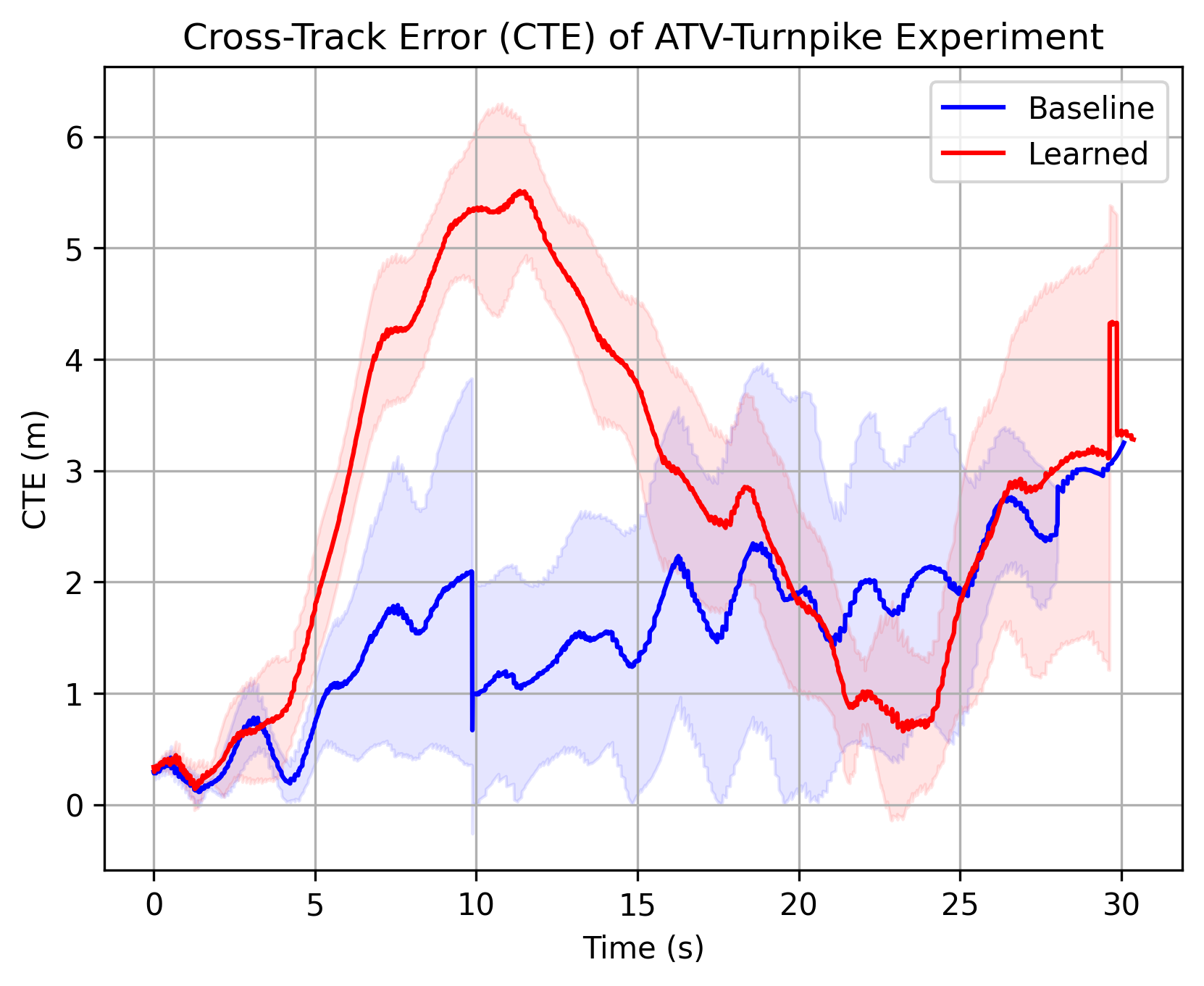}  
  \caption{Average cross-track error between the nominal trajectory and each of the paths taken by the robot over time.}
  \label{fig:turnpike_cte}
\end{subfigure}
\caption{ATV-Pole experiment.}
\label{fig:turnpike}
\end{figure}

\begin{figure}
    \centering
    \includegraphics[width=.8\linewidth]{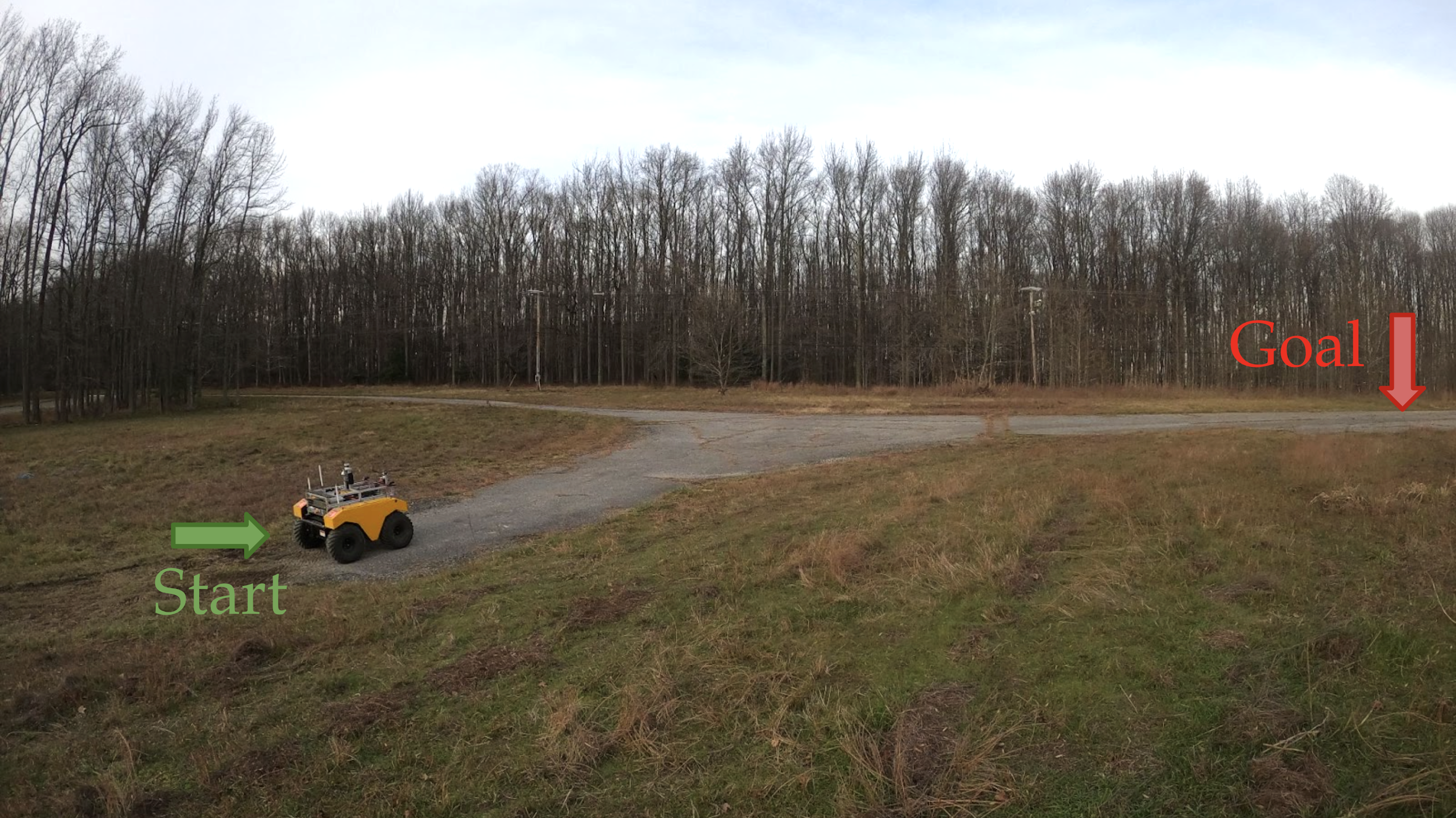}
    \caption{UGV-Hill-Base-Right experiment, where the goal is about 50 m diagonally to the right.}
    \label{fig:warthog_hill_exp}
\end{figure}

\begin{figure}
    \centering
    \includegraphics[width=.8\linewidth]{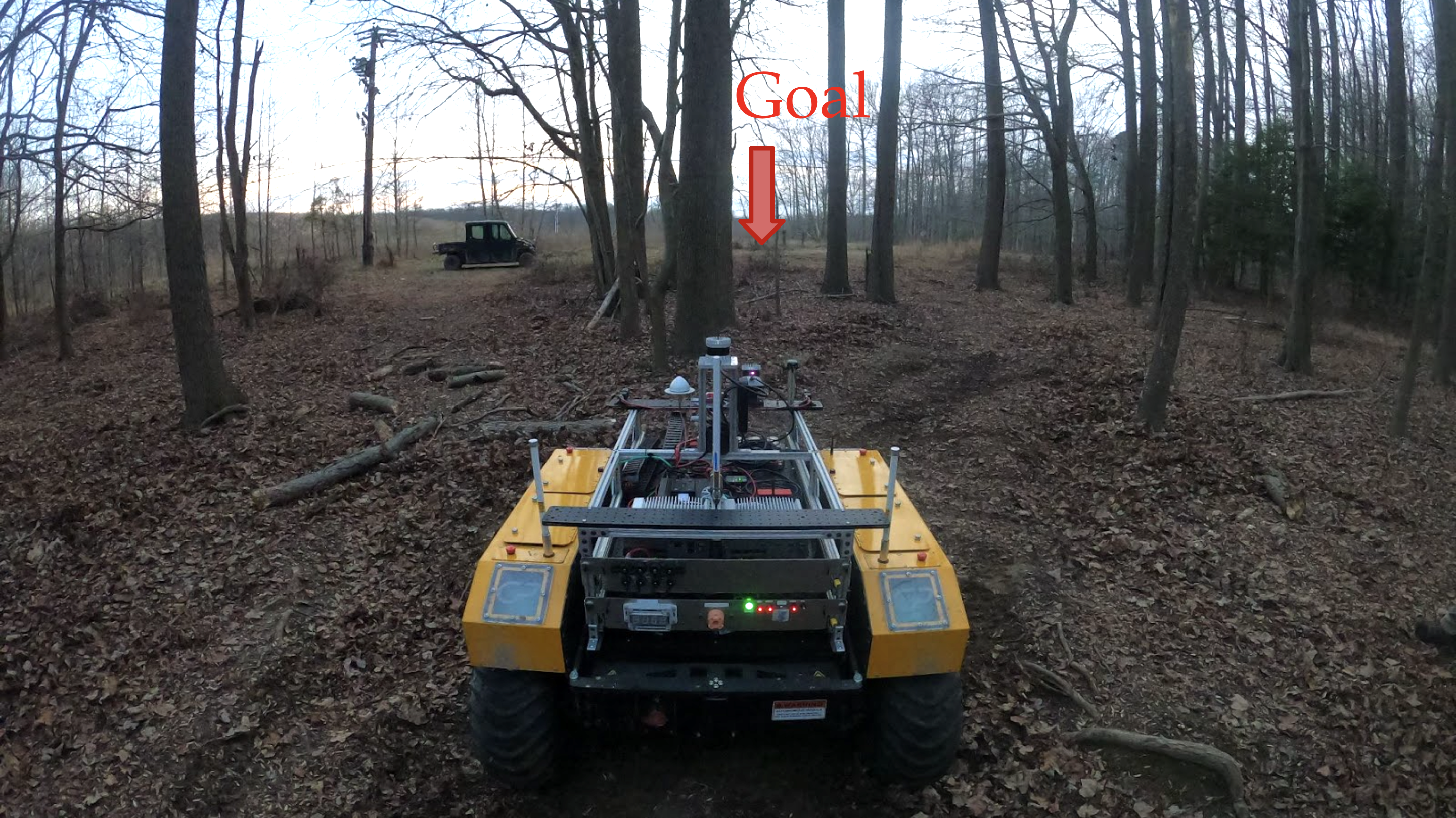}
    \caption{UGV-Forest-Fork experiment, where the goal is straight ahead.}
    \label{fig:warthog_forest_exp}
\end{figure}

We ran two experiments using the ATV robot (ATV-Warehouse shown in Figure \ref{fig:warehuose} and ATV-Pole shown in Figure \ref{fig:turnpike}), and two experiments using the Warthog UGV robot (UGV-Hill-Base-Right shown in Figure \ref{fig:warthog_hill_exp} and UGV-Forest-Fork shown in Figure \ref{fig:warthog_forest_exp}). In this section, we describe the experiments in more detail. Since the experiments are better observed from video, we urge the reader to watch the video in our website.

\textbf{ATV-Warehouse}: In this experiment, there was a patch of grass directly in front of the robot, and there were two smoother dirt paths at each side. As seen in Figure \ref{fig:warehouose_paths}, in most cases, the robot running our learned costmaps took the dirt path to the right of the patch of grass. In one occasion, the robot took the path to the left. While at first this avoided the patch of grass, eventually the robot needed to cut into the patch to get to the goal. This run was stopped early due to an intervention to prevent damaging the robot. When using the baseline costmap, the robot consistently drove straight over the patch of grass. We ran the navigation course five times for each costmap. As observed from the cross track error in Figure \ref{fig:warehouse_cte}, the robot consistently takes a path that strays away from the nominal path to go over smoother terrain.

\textbf{ATV-Pole}: In this experiment, there was also a patch of grass in between the robot and its goal, with two dirt paths surrounding it. However, the dirt paths were also bordered with loose gravel. As seen in the supplemental video, the robot identifies both the grass and the loose gravel as having higher cost than the smooth dirt path, which leads the robot to avoid it. Figure \ref{fig:turnpike_paths} shows the paths that the robot took. One of the baseline runs was stopped early to prevent damage to the robot. We ran the navigation course three times for each costmap.

% \textbf{UGV-Road-to-Hill}: In this experiment, the Warthog robot was given a GPS waypoint as a goal about 50 m diagonally to the right. Given the skid-steer geometry of the robot, the baseline was able to navigate almost straight to the goal. We ran a single trial using the baseline, and three using our learned costmaps. As seen in Figure \ref{fig:warthog_paths}, in all three, the robot takes a longer part that deviates from the nominal straight line path, avoiding the tall grass and instead taking a dirt path to the goal. As discussed in Section 5 of the main paper, this is in part due to the motion constraint imposed by the sight cone of the robot. 

\begin{figure}
    \centering
    \includegraphics[width=.8\linewidth]{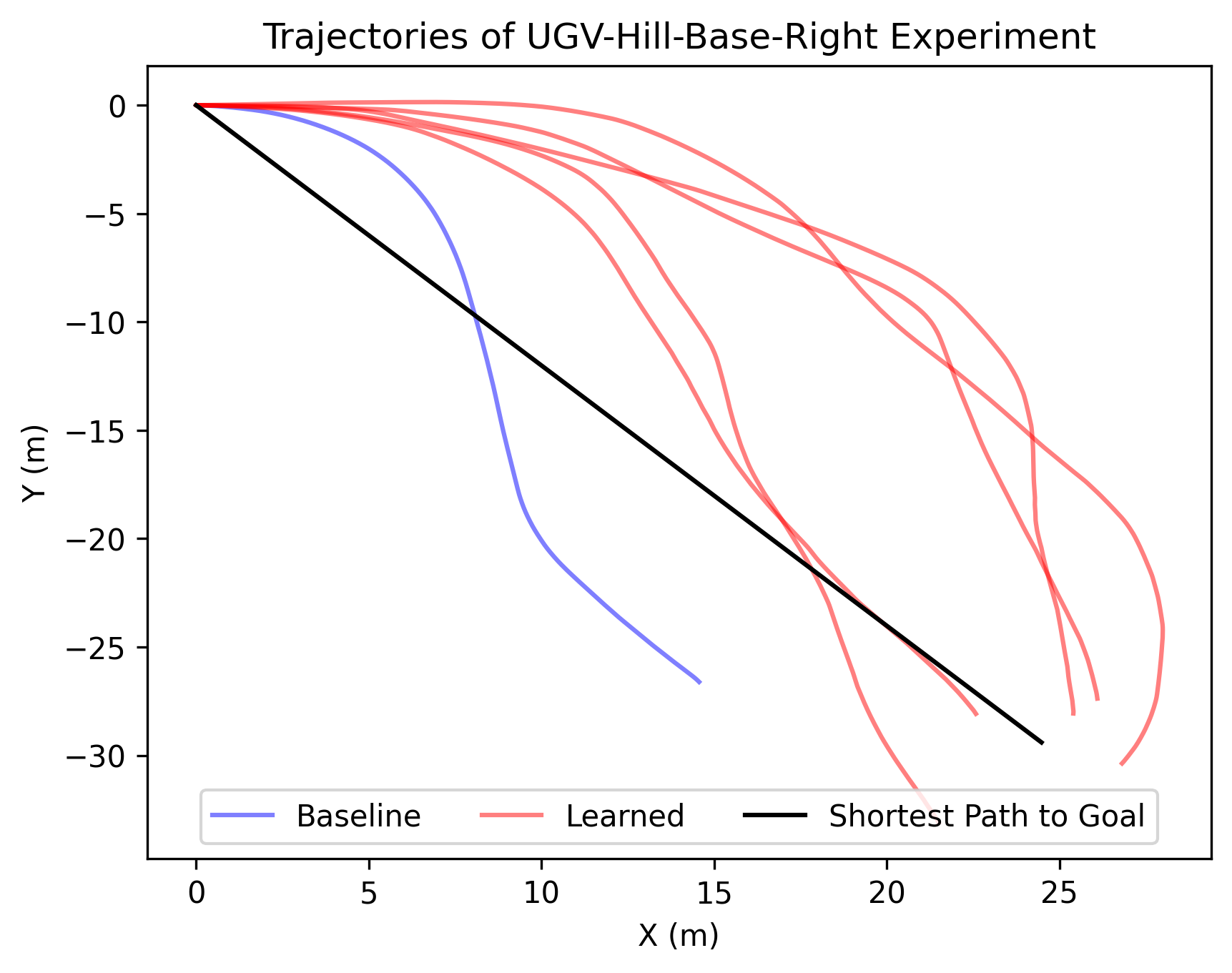}
    \caption{Paths taken by the Warthog robot. Nominal trajectory (black) represents the straight path between the origin and the goal.}
    \label{fig:warthog_hill}
\end{figure}

\textbf{UGV-Hill-Base-Right}: In this experiment (Figure \ref{fig:warthog_hill_exp}), the Warthog robot was given a GPS waypoint as a goal about 50 m diagonally to the right. This goal was located along a concrete path that surrounds a patch of vegetation. The baseline navigation stack immediately starts turning and navigates straight to the goal, going over the vegetation, since it does not reason about the difference in traversability properties between the smooth concrete and the grass. We ran a single trial using the baseline, and five using our learned costmaps. As seen in Figure \ref{fig:warthog_hill}, in all five runs with our costmaps, the robot takes a longer path that deviates from the nominal straight line path, avoiding the grass and instead taking instead the smoother concrete path to the goal. 

\begin{figure}
    \centering
    \includegraphics[width=.8\linewidth]{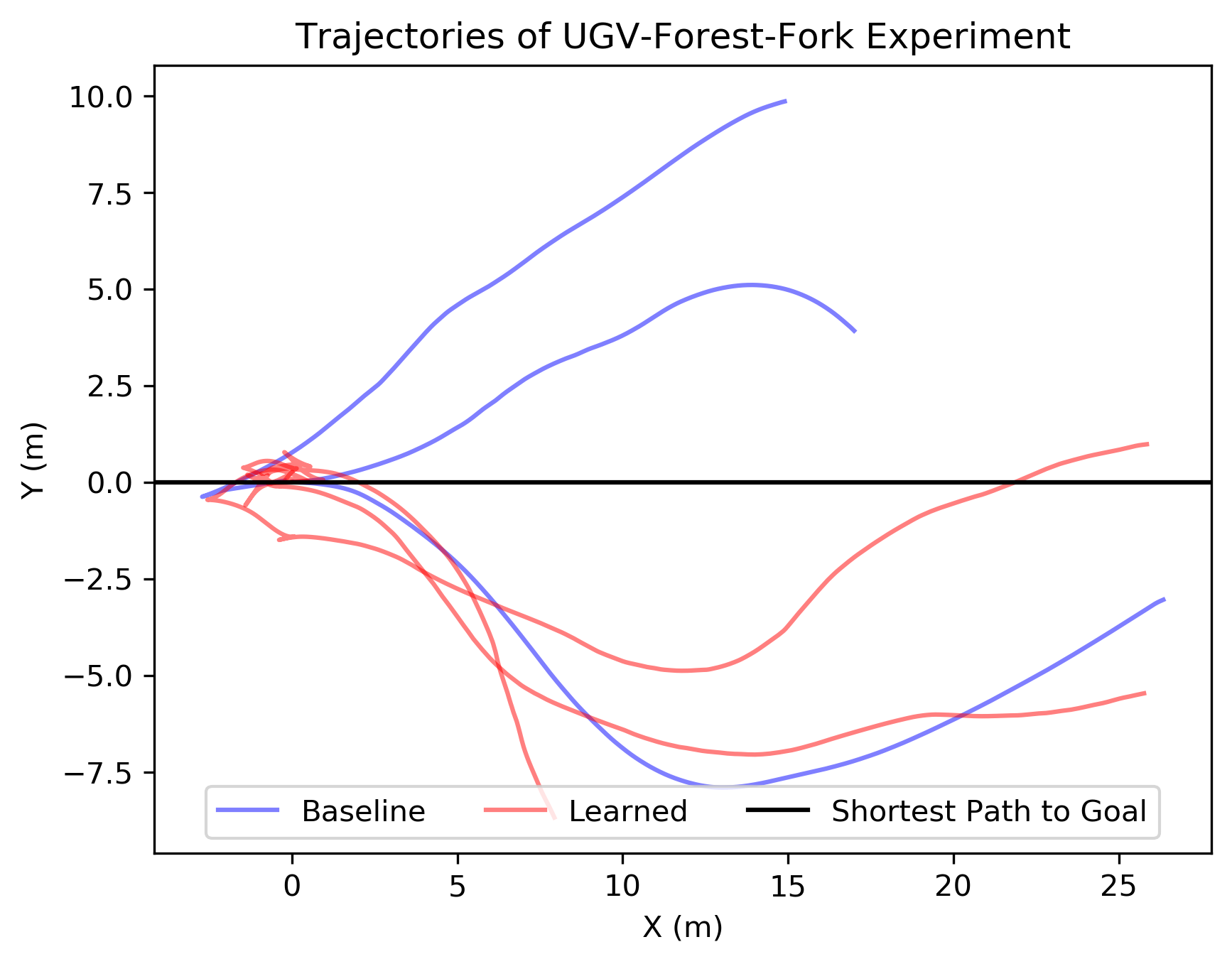}
    \caption{Paths taken by the Warthog robot. Nominal trajectory (black) represents the straight path between the origin and the goal.}
    \label{fig:warthog_forest}
\end{figure}

\textbf{UGV-Forest-Fork}: In this experiment (Figure \ref{fig:warthog_forest_exp}), the Warthog is placed before a fork in a forest trail, where both sides of the fork lead to the same spot. We artificially litter the left path of the fork with tree branches, rocks, and fallen leaves. On the other hand, we clear the path on the right of all obstacles to make it as smooth and possible. Note that none of the small obstacles on the left path would be registered as lethal obstacles using a geometry-based lethal-height binary costmap. Similarly, it is likely that these two trails would belong to the same semantic class using a visual semantic classifier. Therefore, reasoning about fine-grained features in the trails is required for the robot to choose the smoother path. We ran three trials with the baseline stack, and three trials with our costmap. With the baseline stack, the robot takes the left path two times, and the right path once. When using our costmaps, we first move the robot left-to-right in-place manually to fill in the RGB map (since the visual range is shorter than that of the lidar-based baseline stack). We observe that in all three runs, the robot prefers the smoother path to reach the goal. The trajectories are shown in Figure \ref{fig:warthog_forest}.

% \begin{figure}[ht]
% \begin{subfigure}{.45\textwidth}
%   \centering
%   % include first image
%   \includegraphics[width=.8\linewidth]{media/warthog_paths.png}  
%   \caption{Paths taken by the robots. The goal for this experiment was not in a straight line in front of the robot.}
%   \label{fig:warthog_paths}
% \end{subfigure}
% \hfill
% \begin{subfigure}{.45\textwidth}
%   \centering
%   % include second image
%   \includegraphics[width=.8\linewidth]{media/warthog_cte.png}  
%   \caption{Average cross-track error between the shortest line to goal and each of the paths taken by the robot over time.}
%   \label{fig:warthog_cte}
% \end{subfigure}
% \caption{UGV-Road-to-Hill experiment.}
% \label{fig:warthog}
% \end{figure}

\subsubsection{Large-Scale Navigation Experiments:}

We ran experiments on three large courses to compare the performance between the baseline navigation stack (which uses only a lethal height costmap) and our navigation stack (which combines the lethal height costmap with our learned costmaps). In this section, we describe the courses and experiments in more detail. We urge the reader to watch the videos for these experiments in our supplemental material.

\textbf{Red Course}: The red course consists of flat forest trails of finer gravel, with vegetation on the sides of the trails. One of the main challenges of this 400 m trail is that it has a couple of tight turns, which are challenging for both the baseline and our proposed navigation stack.

\textbf{Blue Course}: This 3150 m course consists of flat and hilly terrain, which ranges from smooth gravel to large pebbles, notably in the sloped sections. Additionally, part of the course is covered in vegetation about 1 m tall.

\textbf{Green Course}: The main challenge of this 900 m trail is that about half the trail is covered in tall vegetation.

\subsection{High-Resolution Examples}

\begin{figure}[ht]
    \centering
    \includegraphics[width=\linewidth]{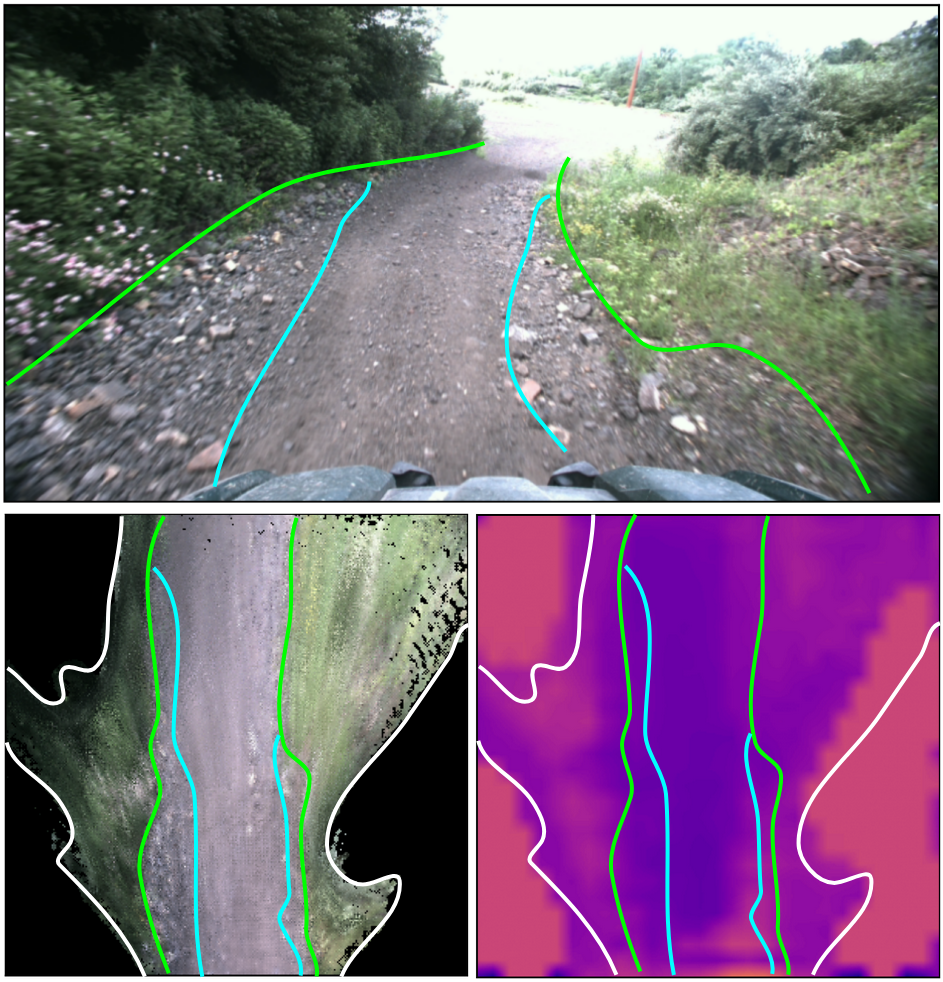}
    \caption{Side-by-side comparison of front-facing view (top row), RGB map (bottom left) and our predicted costmap (bottom right), with approximate hand-annotations separating the different types of terrain. The white lines enclose the invalid regions, the green lines enclose the regions with vegetation, and the blue lines enclose areas with gravel in the terrain. The predicted costmap predicts a higher cost for grass \textit{and} gravel. Notice that gravel appears as a higher-frequency texture in the RGB map.}
    \label{fig:high-res-1}
\end{figure}

\begin{figure}[ht]
    \centering
    \includegraphics[width=\linewidth]{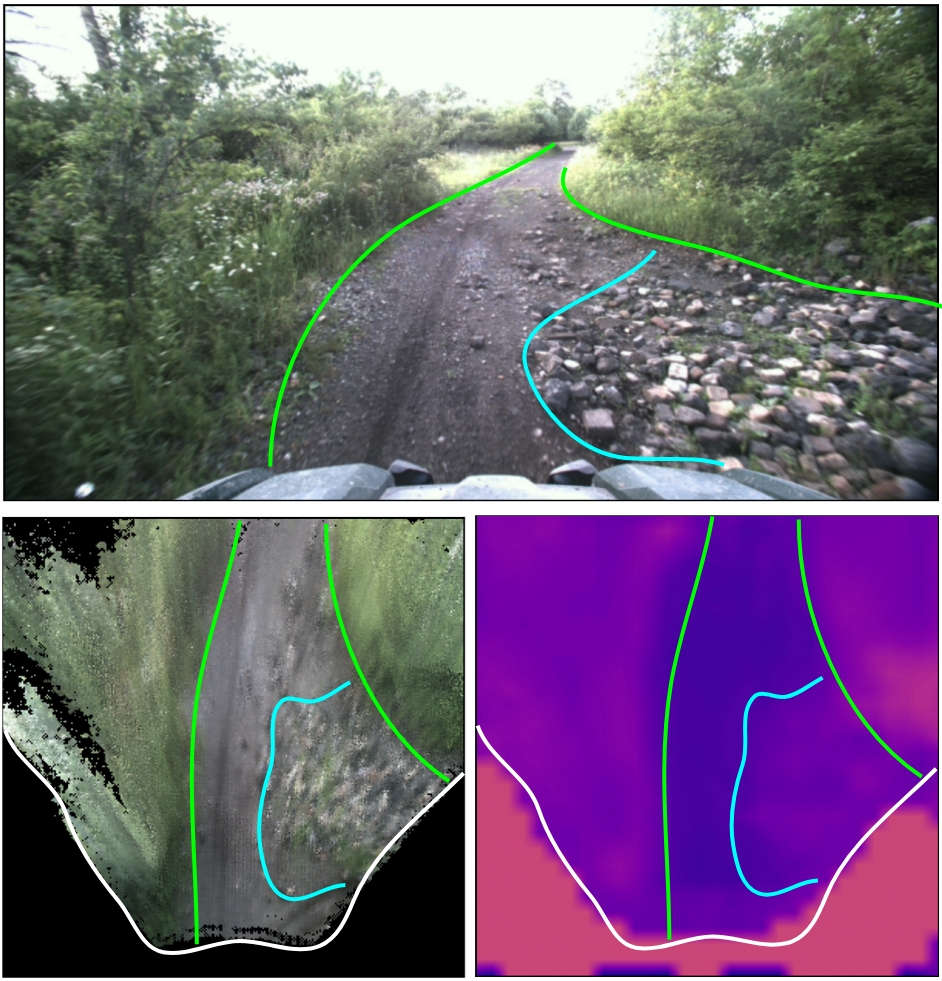}
    \caption{Side-by-side comparison of front-facing view (top row), RGB map (bottom left) and our predicted costmap (bottom right), with approximate hand-annotations separating the different types of terrain. The white lines enclose the invalid regions, the green lines enclose the regions with vegetation, and the blue lines enclose areas with gravel in the terrain. The predicted costmap predicts a higher cost for grass \textit{and} gravel. Notice that gravel appears as a higher-frequency texture in the RGB map.}
    \label{fig:high-res-2}
\end{figure}

In this section, we show two high-resolution examples comparing the top-down RGB map and the predicted costmap side-by-side, in Figures \ref{fig:high-res-1} and \ref{fig:high-res-2}.

\end{document}